\documentclass{article}
\usepackage{iclr2027_conference,times}

\usepackage{amsmath,amsfonts,bm}

\def\eqref#1{equation~\ref{#1}}

\def\1{\bm{1}}

\DeclareMathAlphabet{\mathsfit}{\encodingdefault}{\sfdefault}{m}{sl}
\SetMathAlphabet{\mathsfit}{bold}{\encodingdefault}{\sfdefault}{bx}{n}

\usepackage{url}
\usepackage{graphicx}
\usepackage{float}
\usepackage{capt-of}
\usepackage{wrapfig}
\usepackage{placeins}
\usepackage{multirow}
\usepackage{colortbl}
\usepackage{booktabs}
\usepackage{hyperref}

\definecolor{rankfirst}{RGB}{220,239,210}
\definecolor{ranksecond}{RGB}{255,246,196}
\definecolor{rankthird}{RGB}{207,230,240}
\newcommand{\first}[1]{\cellcolor{rankfirst}#1}
\newcommand{\second}[1]{\cellcolor{ranksecond}#1}
\newcommand{\third}[1]{\cellcolor{rankthird}#1}

\title{SplatAnything: \\ Gaussian Spacing Normalization for Feed-forward 3DGS across Arbitrary Scenes}

\author{%
\textbf{Hao Yu}$^{1,2}$ \quad
\textbf{Zhenwei Wang}$^{2}$ \quad
\textbf{Yongzhen Hu}$^{1}$ \quad
\textbf{Yuning Gong}$^{2}$ \quad
\textbf{Junta Wu}$^{2}$ \\
\textbf{Xiaowei Zhou}$^{1}$ \quad
\textbf{Hujun Bao}$^{1}$ \quad
\textbf{Tengfei Wang}$^{2,\dagger}$ \quad
\textbf{Chunchao Guo}$^{2,\dagger}$ \quad
\textbf{Sida Peng}$^{1,3}$ \\
$^{1}$Zhejiang University \quad
$^{2}$Tencent Hunyuan \quad
$^{3}$Causa Robotics \\
$^{\dagger}$Corresponding author%
}

\iclrfinalcopy

\hypersetup{
  colorlinks=true,
  linkcolor=black,
  citecolor=black,
  urlcolor=blue,
  pdftitle={SplatAnything: Gaussian Spacing Normalization for Feed-forward 3DGS across Arbitrary Scenes},
  pdfauthor={Hao Yu, Zhenwei Wang, Yongzhen Hu, Yuning Gong, Junta Wu, Xiaowei Zhou, Hujun Bao, Tengfei Wang, Chunchao Guo, Sida Peng}
}

\begin{document}
\pagestyle{plain}
\raggedbottom

\maketitle

\vspace{-0.15in}
\begin{figure}[h]
    \centering
    \includegraphics[width=\textwidth]{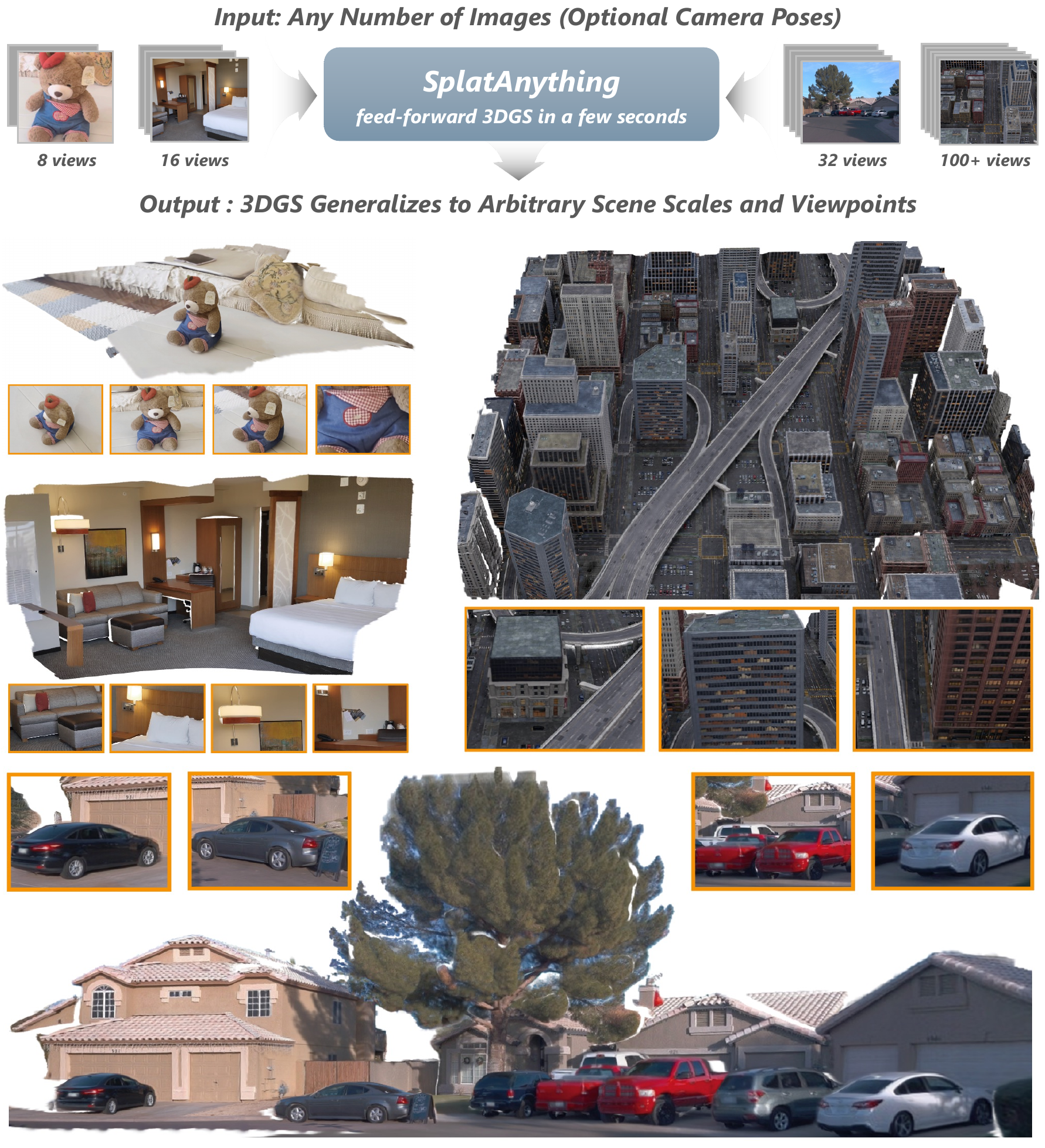}
    \caption{
        Given any number of input images and optional camera poses,
        \textbf{SplatAnything} reconstructs a 3D Gaussian Splatting (3DGS)
        representation within seconds. It generalizes across arbitrary scene
        scales and camera configurations, while supporting challenging
        extrapolated viewpoints. Compared with prior feed-forward 3DGS methods,
        SplatAnything achieves substantially stronger cross-scene generalization
        and novel-view rendering quality.
    }
    \label{fig:teaser}
\end{figure}

\clearpage
\begin{abstract}

    We identify a limitation of feed-forward 3DGS methods: Gaussian attributes are typically predicted in a \textit{radial-distance-normalized} space that canonicalizes a global radial statistic of the reconstructed geometry, while the spacing between neighboring Gaussian centers can still vary substantially across scene scales and camera configurations. This mismatch induces systematic shifts in Gaussian-scale targets, leading to ambiguous supervision.
    To address this, we propose \textbf{SplatAnything} with \textit{Gaussian Spacing Normalization (GSNorm)}, which canonicalizes a robust scene-level statistic of Gaussian spacing to provide a more consistent spatial reference for Gaussian prediction. To rigorously assess our method, we further establish a systematic benchmark spanning diverse datasets, input sparsity levels, and challenging extrapolated viewpoints. Extensive experiments demonstrate that SplatAnything achieves state-of-the-art performance, with particularly strong gains under substantial variations in scene scale and viewpoint.
    \href{https://zju3dv.github.io/SplatAnything/}{Project page}.

\end{abstract}
\section{Introduction}
\label{sec:introduction}


Feed-forward 3D Gaussian Splatting (3DGS) directly predicts a 3D
Gaussian representation~\citep{kerbl20233d} from input images, avoiding
the costly per-scene optimization of conventional 3DGS.
This efficiency makes it attractive for large-scale 3D reconstruction and
digital content creation.
The ultimate goal is a single model that generalizes reliably across
arbitrary scene scales, camera configurations, and target viewpoints, yet
such broad generalization remains challenging.

A prevailing paradigm is to first predict per-pixel depth using a
geometry foundation model~\citep{wang2025vggt,lin2025depth,team2026hy,jiang2026omnix,yu2026infinidepth},
and then predict per-pixel Gaussian attributes based on the resulting
geometry in its \emph{radial-distance-normalized} coordinate space.
\emph{Radial Distance Normalization (RDNorm)} uses the mean radial
distance of reconstructed 3D points to the reference-camera origin as a
scene-level scale, and globally rescales the scene to canonicalize this
statistic. However, RDNorm \textbf{does not} directly canonicalize the spacing between
neighboring Gaussian centers, which we refer to as \emph{Gaussian
spacing}. Since adequate surface coverage requires the Gaussian scale to match the
local Gaussian spacing, variations in this quantity directly induce shifts
in the corresponding learning targets.

\begin{figure}[t]
    \centering
    \includegraphics[width=\textwidth]{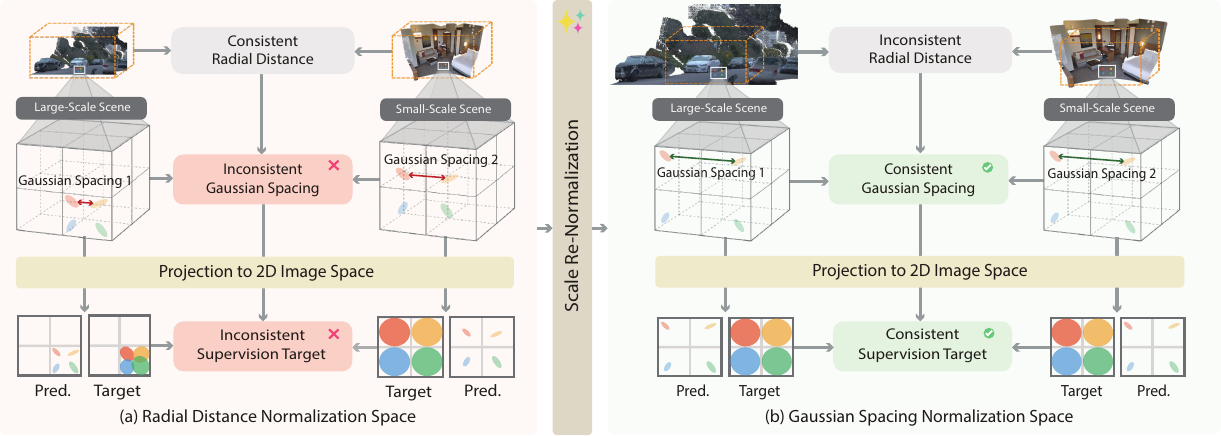}
    \vspace{-0.2in}
    \caption{
        \textbf{Normalization space.}
        (a) \emph{Radial Distance Normalization} canonicalizes a global
        radial statistic but leaves Gaussian spacing varying across scenes
        and camera configurations, resulting in inconsistent
        supervision targets.
        (b) \emph{Gaussian Spacing Normalization} canonicalizes a representative statistic of Gaussian spacing, providing a more consistent spatial reference for Gaussian prediction.
    }
    \vspace{-0.2in}
    \label{fig:gsnorm}
\end{figure}

As illustrated in Fig.~\ref{fig:gsnorm}, visually similar regions can exhibit
substantially different Gaussian spacings after RDNorm under different
scene scales and camera configurations. Consequently, similar local
observations may correspond to different Gaussian-scale targets across
scenes, introducing ambiguity in scale supervision.
A predictor biased toward smaller Gaussian scales may
under-cover regions with large Gaussian spacing and produce holes,
whereas one biased toward larger scales may cause excessive overlap in
densely sampled regions and lead to over-smoothing, as shown in
Fig.~\ref{fig:ablation}.
This ambiguity in Gaussian-scale supervision poses a major bottleneck to
robust feed-forward 3DGS generalization across varying scene scales and
camera configurations.

To address this issue, we introduce \textbf{SplatAnything}, a
feed-forward 3DGS framework equipped with \emph{Gaussian Spacing
Normalization (GSNorm)}. The key idea of GSNorm is to derive the
scene-level normalization factor from a representative statistic of
Gaussian spacing, rather than from a global radial statistic. Under the
per-pixel Gaussian prediction paradigm, where each valid image pixel
corresponds to a Gaussian center, perspective projection relates Gaussian
spacing to scene depth and camera intrinsics. Based on this relation,
GSNorm computes Gaussian spacing over all valid observations and uses its
median as a robust scene-level normalization factor. The reconstructed
geometry is then rescaled accordingly, canonicalizing the representative
Gaussian spacing across scenes. This provides a more consistent spatial reference for Gaussian-scale prediction, alleviating ambiguity in supervision caused by variations in
scene scale and camera configuration and thereby improving generalization
across diverse scenes and viewpoints.

To rigorously assess SplatAnything and prior methods, we further
introduce a cross-dataset benchmark for in-the-wild evaluation.
Unlike conventional protocols~\citep{wang2026tttlrm,xu2025depthsplat,
xu2025resplat,ye2026yonosplat,ren2026tokengs}, which often focus on
training-related domains and interpolated target views, our benchmark
covers diverse datasets, scene scales, image resolutions, and camera
configurations, while explicitly varying input-view sparsity and
emphasizing challenging extrapolated viewpoints. Extensive experiments show that SplatAnything
achieves state-of-the-art performance, with particularly strong gains under large scene and viewpoint shifts.

In summary, our contributions are:

\vspace{-0.05in}

\begin{itemize}
    \item
    We identify ambiguity in Gaussian-scale supervision across varying
    scene scales and camera configurations induced by RDNorm. To address
    this issue, we propose \emph{Gaussian Spacing Normalization (GSNorm)},
    which canonicalizes a robust scene-level statistic of Gaussian spacing
    to provide a more consistent spatial reference for Gaussian prediction.

    \item
    We establish a systematic cross-dataset evaluation benchmark for
    feed-forward 3DGS, spanning diverse scene categories, image
    resolutions, camera configurations, input-view sparsity levels, and
    challenging extrapolated viewpoints. This benchmark enables a
    controlled evaluation of cross-scene and cross-view generalization.
\end{itemize}
\section{Related Work}
\label{sec:related-work}
\paragraph{Geometry foundation models.}
Recent feed-forward geometry foundation models~\citep{wang2024dust3r,wang2025vggt,lin2025depth,liu2025worldmirror,jiang2026omnix} directly recover depth, point maps, and camera poses from RGB observations, providing strong geometric priors across diverse scenes.
These models typically employ radial distance normalization to keep reconstructed geometry within a numerically stable range. While effective for geometry prediction, such normalization does not ensure consistent \emph{Gaussian spacing} across scenes, cameras, and image resolutions. This becomes critical when the normalized geometry is reused for per-pixel Gaussian prediction, since Gaussian scales are defined in the same 3D coordinate system.
\paragraph{Feed-forward 3D Gaussian Splatting.}
Feed-forward 3DGS methods directly infer Gaussian representations from image observations, often by first estimating per-pixel depth and then predicting Gaussian attributes. pixelSplat~\citep{charatan2024pixelsplat} reconstructs 3D Gaussians from image pairs by learning a 3D probability distribution over Gaussian centers. DepthSplat~\citep{xu2025depthsplat} combines monocular depth priors with multi-view matching, while ReSplat~\citep{xu2025resplat} further refines predicted Gaussians through rendering-based optimization. However, limited geometric accuracy can become a bottleneck under large viewpoint extrapolation.
More recent methods~\citep{jiang2025anysplat,lin2025depth,team2026hy,
anonymous2027infinisplatv2} leverage strong geometry foundation models
to initialize Gaussian centers and predict the remaining attributes.
Although this substantially improves reconstruction quality, these
methods still inherit the radial-distance-normalized space of their
geometry backbones, in which Gaussian spacing can vary substantially
across scene scales and camera configurations, leading to ambiguity in
Gaussian-scale supervision. Such supervision ambiguity makes it difficult
to learn a single Gaussian predictor that generalizes reliably across
diverse input settings.
Other approaches~\citep{an2025c3g,li2026querysplat} do not rely on
explicit geometric priors. While they can achieve strong rendering
quality at interpolated viewpoints, the learned Gaussian geometry can
become less reliable under large viewpoint extrapolation.

\section{SplatAnything}
\label{sec:method}

\subsection{Formulation}

Given $N$ input images
$\mathcal{I}=\{I_i\}_{i=1}^{N}$,
together with optional camera intrinsics
$\mathcal{K}=\{K_i\}_{i=1}^{N}$
and extrinsics
$\mathcal{T}=\{(R_i,t_i)\}_{i=1}^{N}$,
our goal is to predict a 3D Gaussian scene representation
\begin{equation}
    \mathcal{G}
    =
    \left\{
        \left(\mu_j,\alpha_j,\Sigma_j,\mathrm{sh}_j\right)
    \right\}_{j=1}^{M},
\end{equation}
where $\mu_j$, $\alpha_j$, $\Sigma_j$, and $\mathrm{sh}_j$
denotethe Gaussian center, opacity, covariance, and SH coefficients.

\subsection{Preliminaries}

\paragraph{Gaussian spacing.}
For per-pixel Gaussian prediction, each valid image pixel defines a
Gaussian center at its back-projected 3D point. We refer to the spatial
spacing between Gaussian centers induced by neighboring image pixels as
\emph{Gaussian spacing}. This quantity is particularly relevant to
Gaussian-scale prediction, since adequate surface coverage requires the
Gaussian scale to be commensurate with the local Gaussian spacing.

For the $i$-th view, let $D_i(\mathbf{u})$ denote the depth at pixel
$\mathbf{u}=(u,v)$, and let $(f_{x,i},f_{y,i})$ and
$(c_{x,i},c_{y,i})$ denote the focal lengths and principal point in
pixels. Under the pinhole camera model, the back-projected point $\mathbf{P}_i^{c}(\mathbf{u})$ in the
local camera frame is
\begin{equation}
    \mathbf{P}_i^{c}(\mathbf{u})
    =
    D_i(\mathbf{u})
    \begin{bmatrix}
        (u-c_{x,i})/f_{x,i} \\
        (v-c_{y,i})/f_{y,i} \\
        1
    \end{bmatrix}.
    \label{eq:back_projection}
\end{equation}

Consider two horizontally adjacent pixels $\mathbf{u}$ and
$\mathbf{u}+\mathbf{e}_x$. Under a locally constant-depth approximation,
$D_i(\mathbf{u}+\mathbf{e}_x)\approx D_i(\mathbf{u})$, their horizontal
Gaussian spacing is
\begin{equation}
    d_{x,i}(\mathbf{u})
    =
    \left\|
        \mathbf{P}_i^{c}(\mathbf{u}+\mathbf{e}_x)
        -
        \mathbf{P}_i^{c}(\mathbf{u})
    \right\|_2
    \approx
    \frac{D_i(\mathbf{u})}{f_{x,i}}.
    \label{eq:horizontal_gaussian_spacing}
\end{equation}
Similarly, the vertical spacing is
$d_{y,i}(\mathbf{u})\approx D_i(\mathbf{u})/f_{y,i}$.
We summarize the two directions using their geometric mean,
\begin{equation}
    d_i(\mathbf{u})
    =
    \sqrt{
        d_{x,i}(\mathbf{u})d_{y,i}(\mathbf{u})
    }
    \approx
    \frac{D_i(\mathbf{u})}
    {\sqrt{f_{x,i}f_{y,i}}},
    \label{eq:gaussian_spacing}
\end{equation}
which we use as a first-order scalar characterization of local Gaussian
spacing.

\subsection{Normalization Space for Gaussian Prediction}
\label{sec:gsnorm}

\paragraph{Limitation of radial distance normalization.}

Feed-forward 3DGS methods built on geometry foundation models typically
predict Gaussian attributes in the normalized coordinate space inherited
from their geometry backbones. These models commonly normalize a reconstructed scene using the mean radial distance of 3D points to the reference-camera origin, which we
refer to as \emph{Radial Distance Normalization (RDNorm)}.
Let $\mathbf{P}_i(\mathbf{u})$ denote the reconstructed 3D point at
pixel $\mathbf{u}$ in view $i$, expressed in the reference-camera
coordinate system before scene-level normalization, and let $\Omega_i$
denote the set of valid pixels in view $i$.
The normalization factor $s_r$ under RDNorm is computed as
\begin{equation}
    s_r
    =
    \frac{1}{N_p}
    \sum_{i=1}^{N}
    \sum_{\mathbf{u}\in\Omega_i}
    \left\|
        \mathbf{P}_i(\mathbf{u})
    \right\|_2,
    \qquad
    N_p
    =
    \sum_{i=1}^{N}|\Omega_i|.
    \label{eq:radial_distance}
\end{equation}

\begin{figure}[t]
    \centering
    \includegraphics[width=\textwidth]{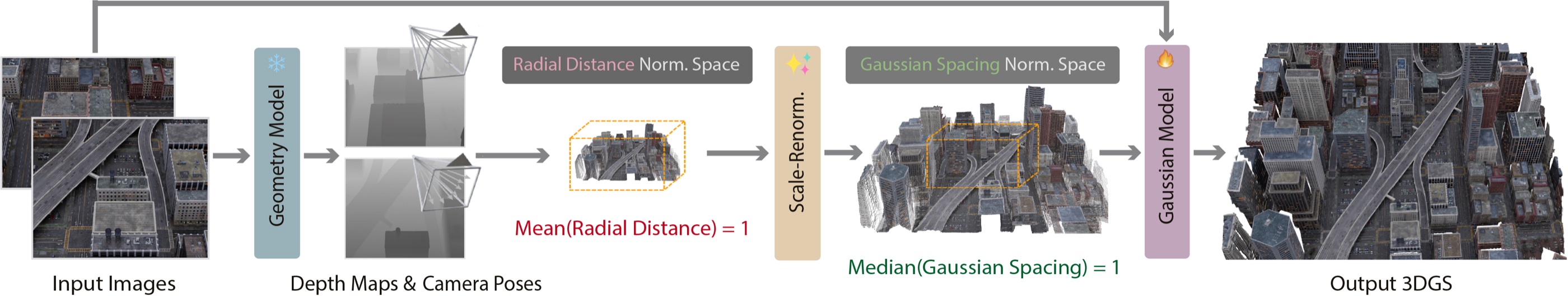}
    \vspace{-0.2in}
    \caption{
        \textbf{Architecture of SplatAnything.}
        Given $N$ input images and optional camera parameters,
        a frozen geometry foundation model first estimates multi-view
        geometry.
        We convert the geometry from its native radial-distance-normalized
        space to the proposed Gaussian-spacing-normalized space and predict
        per-pixel Gaussian attribute through a Gaussian backbone and a lightweight GShead.
    }
    \label{fig:architecture}
    \vspace{-0.12in}
\end{figure}

The geometry, depth, and camera translations are then rescaled as
\begin{equation}
    \mathbf{P}_i^{r}
    =
    \frac{\mathbf{P}_i}{s_r},
    \qquad
    D_i^{r}
    =
    \frac{D_i}{s_r},
    \qquad
    \mathbf{t}_i^{r}
    =
    \frac{\mathbf{t}_i}{s_r},
    \label{eq:rdnorm}
\end{equation}

From Eq.~\ref{eq:gaussian_spacing}, the Gaussian spacing after RDNorm is
approximately
\begin{equation}
    d_i^{r}(\mathbf{u})
    \approx
    \frac{D_i^{r}(\mathbf{u})}
    {\sqrt{f_{x,i}f_{y,i}}}
    =
    \frac{D_i(\mathbf{u})}
    {\sqrt{f_{x,i}f_{y,i}}\,s_r}.
    \label{eq:rdnorm_spacing}
\end{equation}

RDNorm canonicalizes a scene-level radial statistic of the reconstructed geometry, but does not directly canonicalize a scene-level statistic of Gaussian spacing.

\begin{wrapfigure}[16]{r}{0.42\linewidth}
    \centering
    \includegraphics[width=\linewidth]{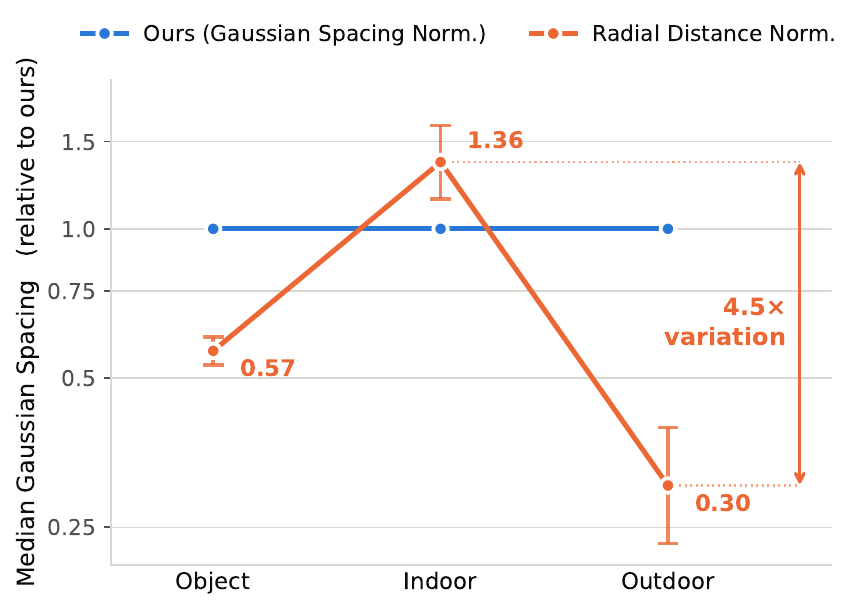}
    \vspace{-0.25in}
    \caption{
        \textbf{Median Gaussian spacing across scene types.}
        RDNorm leaves up to $4.5\times$ variation in median Gaussian spacing
        across scene categories, whereas GSNorm canonicalizes it by
        construction.
    }
    \vspace{-0.1in}
    \label{fig:point_spacing_comparison}
\end{wrapfigure}

Consequently, the spacing between neighboring Gaussian centers can still
vary substantially across different scene scales and camera
configurations, as shown in
Fig.~\ref{fig:point_spacing_comparison}.
This variation is particularly
important for Gaussian prediction because the Gaussian scale required
for adequate surface coverage should be commensurate with the local
Gaussian spacing. Therefore, similar local observations can correspond
to different Gaussian-scale targets after RDNorm,
introducing ambiguity in scale supervision across scenes. This motivates us to
define the normalization space directly from a representative statistic
of Gaussian spacing.

\paragraph{Gaussian spacing normalization.}
We propose \emph{Gaussian Spacing Normalization (GSNorm)}, which
defines the coordinate scale for Gaussian prediction using Gaussian spacing
rather than a global radial statistic.
Because the reconstruction is rescaled by a single scene-level factor,
we use the median characteristic Gaussian spacing as a robust representative
scale.
Specifically, for a generic scene-level scale $s>0$, the characteristic
Gaussian spacing becomes
\begin{equation}
    d_i^{s}(\mathbf{u})
    \approx
    \frac{D_i(\mathbf{u})}
    {\sqrt{f_{x,i}f_{y,i}}\,s}.
\end{equation}
GSNorm chooses $s$ such that the scene-level median Gaussian spacing is
canonicalized:
\begin{equation}
    \operatorname*{median}_{i,\mathbf{u}\in\Omega_i}
    \left(
        d_i^{s}(\mathbf{u})
    \right)
    \approx
    \operatorname*{median}_{i,\mathbf{u}\in\Omega_i}
    \left(
        \frac{D_i(\mathbf{u})}
        {\sqrt{f_{x,i}f_{y,i}}\,s}
    \right)
    =
    1.
    \label{eq:gsnorm_objective}
\end{equation}

Since the median is equivariant to positive scaling,
Eq.~\ref{eq:gsnorm_objective} directly yields the normalization factor $s_g$:
\begin{equation}
    s_g
    =
    \operatorname*{median}_{i,\mathbf{u}\in\Omega_i}
    \left(
        \frac{D_i(\mathbf{u})}
        {\sqrt{f_{x,i}f_{y,i}}}
    \right).
    \label{eq:gs_scale}
\end{equation}

Using $s_g$, we rescale the geometry, depth, and camera translations as
\begin{equation}
    \mathbf{P}_i^{g}
    =
    \frac{\mathbf{P}_i}{s_g},
    \qquad
    D_i^{g}
    =
    \frac{D_i}{s_g},
    \qquad
    \mathbf{t}_i^{g}
    =
    \frac{\mathbf{t}_i}{s_g}.
    \label{eq:gsnorm}
\end{equation}

GSNorm canonicalizes the scene-level median Gaussian spacing while preserving
the relative variation of local Gaussian spacing.
This aligns the representative Gaussian spacing scale across scenes and camera
configurations without suppressing meaningful local geometric variation.
Compared with RDNorm, GSNorm therefore provides a more consistent scale
reference for Gaussian prediction that is directly related to Gaussian
surface coverage.
This alleviates ambiguity in Gaussian-scale supervision across different
input settings, facilitating the learning of a feed-forward Gaussian
model with robust cross-scene and cross-view generalization.

\subsection{SplatAnything Architecture}
\label{sec:architecture}
The overall architecture is illustrated in
Fig.~\ref{fig:architecture}. More details are provided in the appendix.


\paragraph{Geometry prediction and rescaling.}

Given $N$ input images and optional camera parameters, we employ the
frozen WorldMirror~2.0~\citep{team2026hy} geometry model to estimate
multi-view depths, intrinsics, and camera poses. The predicted depths
and camera translations are initially represented in the
radial-distance-normalized coordinate space of the geometry model.
We then use GSNorm (Eq.~\ref{eq:gsnorm}) to rescale these
scale-dependent quantities into the proposed Gaussian-spacing-normalized
space, obtaining
$D^{g}=\{D_i^{g}\}_{i=1}^{N}$ and
$T^{g}=\{(R_i,t_i^{g})\}_{i=1}^{N}$.
We unproject the rescaled depth maps using the corresponding camera
parameters to obtain
$\mathbf{P}^{g}=\{\mathbf{P}_i^{g}\}_{i=1}^{N}$,
where each valid per-pixel 3D point serves as the center of a Gaussian
primitive.

\paragraph{Gaussian attribute prediction.}

We initialize a \emph{Gaussian Backbone} from the geometry backbone
of WorldMirror~2.0 and fine-tune it for Gaussian prediction.
A lightweight \emph{GSHead}, trained from scratch, predicts
per-pixel Gaussian attributes for all input views:
\begin{equation}
    \{\mathcal{G}_i^{g}\}_{i=1}^{N}
    =
    \mathrm{GSHead}
    \left(
        \mathrm{GaussianBackbone}(\mathcal{I}, Optional\mathcal{(K, T)})
    \right).
\end{equation}

For each pixel, the network predicts opacity, rotation, scale,
and spherical-harmonic appearance coefficients.
Together with the corresponding center from $\mathbf{P}^{p}$,
these quantities define the per-pixel Gaussian primitives in the
Gaussian-spacing-normalized coordinate space.




\subsection{Training SplatAnything}
\label{sec:training}
The training datasets and view-sampling strategies are detailed in the
appendix.

Given a rendered target-view image $\hat{I}_t$ and its ground-truth
image $I_t$, we supervise the model using a pixel-wise MSE loss and an
LPIPS perceptual loss:
\begin{equation}
    \mathcal{L}_{\mathrm{MSE}}
    =
    \frac{1}{3HW}
    \left\|
        \hat{I}_t-I_t
    \right\|_2^2,
    \qquad
    \mathcal{L}_{\mathrm{LPIPS}}
    =
    \operatorname{LPIPS}
    \left(
        \hat{I}_t,I_t
    \right).
\end{equation}

The overall objective is
\begin{equation}
    \mathcal{L}
    =
    \lambda_{\mathrm{MSE}}
    \mathcal{L}_{\mathrm{MSE}}
    +
    \lambda_{\mathrm{LPIPS}}
    \mathcal{L}_{\mathrm{LPIPS}},
\end{equation}
where
$\lambda_{\mathrm{MSE}}=\lambda_{\mathrm{LPIPS}}=1$
in our implementation.

SplatAnything is trained for 20,000 iterations on 64 NVIDIA H20 GPUs,
using 4 to 32 input views and dynamically varying the input resolution
from 504 to 1024. We use a 500-step warm-up and learning rates of
$5\times10^{-6}/2\times10^{-5}$ for the
Gaussian Backbone and GSHead.
\section{Experiments}
\label{sec:experiments}
\begin{table*}[t]
\vspace{-0.15in}
\centering
\caption{Quantitative comparison across seven benchmark datasets, averaged over \textbf{4--64 input views}. The top-3 results are highlighted as \colorbox{rankfirst}{first}, \colorbox{ranksecond}{second}, and \colorbox{rankthird}{third}.}
\vspace{0.03in}
\label{tab:main_comparison_4_64view}
\setlength{\tabcolsep}{1pt}
\renewcommand{\arraystretch}{1.28}
\small
\makebox[\textwidth][c]{%
\resizebox{\textwidth}{!}{%
\begin{tabular}{clcccccccccccccc}
    \toprule
    \multirow{2}{*}{\textbf{Input}}
    & \multirow{2}{*}{\textbf{Method}}
    & \multicolumn{2}{c}{\textbf{ScanNet}}
    & \multicolumn{2}{c}{\textbf{NRGBD}}
    & \multicolumn{2}{c}{\textbf{DL3DV}}
    & \multicolumn{2}{c}{\textbf{Tanks \& Temples}}
    & \multicolumn{2}{c}{\textbf{Mip-NeRF 360}}
    & \multicolumn{2}{c}{\textbf{ETH3D}}
    & \multicolumn{2}{c}{\textbf{DTU}} \\
    \cmidrule(lr){3-4}
    \cmidrule(lr){5-6}
    \cmidrule(lr){7-8}
    \cmidrule(lr){9-10}
    \cmidrule(lr){11-12}
    \cmidrule(lr){13-14}
    \cmidrule(lr){15-16}
    & & DISTS$\downarrow$ & LPIPS$\downarrow$
    & DISTS$\downarrow$ & LPIPS$\downarrow$
    & DISTS$\downarrow$ & LPIPS$\downarrow$
    & DISTS$\downarrow$ & LPIPS$\downarrow$
    & DISTS$\downarrow$ & LPIPS$\downarrow$
    & DISTS$\downarrow$ & LPIPS$\downarrow$
    & DISTS$\downarrow$ & LPIPS$\downarrow$ \\
    \midrule
    \multirow{6}{*}{\rotatebox[origin=c]{90}{w/o Pose}}
    & AnySplat
    & 0.278 & 0.414
    & 0.236 & 0.352
    & 0.371 & 0.327
    & 0.593 & 0.737
    & 0.538 & 0.787
    & 0.278 & 0.154
    & 0.520 & 0.810 \\
    & YoNoSplat
    & 0.265 & \third{0.236}
    & 0.285 & 0.240
    & 0.375 & \third{0.219}
    & 0.424 & \second{0.338}
    & 0.405 & 0.586
    & 0.321 & \second{0.033}
    & 0.484 & 0.729 \\
    & DA3-GS
    & 0.286 & 0.305
    & \second{0.233} & \third{0.217}
    & \third{0.337} & 0.228
    & \third{0.414} & 0.380
    & \third{0.371} & \third{0.565}
    & \third{0.228} & 0.087
    & \third{0.365} & \second{0.582} \\
    & Mirror2-GS
    & \second{0.193} & \second{0.222}
    & \third{0.235} & 0.237
    & \second{0.246} & \second{0.188}
    & \second{0.306} & \third{0.351}
    & \second{0.283} & \second{0.528}
    & \second{0.205} & 0.070
    & \second{0.347} & \third{0.617} \\
    & QuerySplat
    & \third{0.236} & 0.242
    & 0.261 & \second{0.216}
    & 0.343 & 0.291
    & 0.425 & 0.393
    & 0.412 & 0.652
    & 0.290 & \first{0.031}
    & 0.424 & 0.676 \\
    & \textbf{Ours}
    & \first{0.184} & \first{0.207}
    & \first{0.195} & \first{0.183}
    & \first{0.221} & \first{0.156}
    & \first{0.252} & \first{0.299}
    & \first{0.233} & \first{0.463}
    & \first{0.147} & \third{0.049}
    & \first{0.229} & \first{0.483} \\
    \midrule
    \multirow{7}{*}{\rotatebox[origin=c]{90}{w/ Pose}}
    & DepthSplat
    & 0.288 & 0.278
    & 0.323 & 0.272
    & 0.329 & 0.191
    & 0.397 & 0.353
    & 0.561 & 0.656
    & 0.445 & 0.068
    & 0.370 & 0.598 \\
    & TokenGS
    & 0.369 & 0.272
    & 0.333 & 0.277
    & 0.375 & 0.211
    & 0.515 & 0.346
    & 0.502 & 0.716
    & 0.400 & \third{0.042}
    & 0.593 & 0.828 \\
    & tttLRM
    & 0.292 & 0.300
    & 0.320 & 0.292
    & 0.323 & 0.196
    & \third{0.390} & \third{0.342}
    & 0.350 & 0.531
    & 0.225 & 0.063
    & 0.382 & 0.599 \\
    & ReSplat
    & \third{0.272} & \third{0.240}
    & 0.284 & 0.232
    & \third{0.304} & \second{0.159}
    & 0.396 & \first{0.275}
    & \third{0.349} & \third{0.525}
    & \third{0.219} & \first{0.027}
    & 0.441 & 0.664 \\
    & DA3-GS (Pose Cond.)
    & 0.274 & 0.291
    & \third{0.238} & \third{0.227}
    & 0.311 & 0.219
    & 0.425 & 0.368
    & 0.362 & 0.556
    & 0.223 & 0.089
    & \third{0.368} & \third{0.596} \\
    & Mirror2-GS (Pose Cond.)
    & \second{0.195} & \second{0.223}
    & \second{0.231} & \second{0.226}
    & \second{0.239} & \third{0.180}
    & \second{0.304} & 0.346
    & \second{0.272} & \second{0.506}
    & \second{0.201} & 0.062
    & \second{0.315} & \second{0.532} \\
    & \textbf{Ours (Pose Cond.)}
    & \first{0.185} & \first{0.206}
    & \first{0.186} & \first{0.162}
    & \first{0.215} & \first{0.148}
    & \first{0.251} & \second{0.290}
    & \first{0.219} & \first{0.440}
    & \first{0.133} & \second{0.037}
    & \first{0.184} & \first{0.364} \\
    \bottomrule
\end{tabular}%
}%
}
\vspace{-0.05in}

\end{table*}

\subsection{Evaluation Benchmark}
Existing benchmarks are typically concentrated on a small set of datasets and evaluate interpolated target views, making it difficult to assess cross-domain and viewpoint generalization. To provide a more comprehensive evaluation, we design a new benchmark with the following properties:


\textbf{Cross-domain and cross-scale coverage.}
We evaluate on \textbf{seven datasets}: ScanNet~\citep{dai2017scannet}, NRGBD~\citep{azinovic2022neural}, DL3DV~\citep{ling2024dl3dv}, Tanks and Temples~\citep{knapitsch2017tanks}, Mip-NeRF 360~\citep{barron2022mip}, ETH3D~\citep{schops2017multi}, and DTU~\citep{jensen2014large}. These datasets span a wide range of physical scene scales, providing a substantially broader test of cross-domain and cross-scale generalization.


\textbf{Controlled extrapolation and observation sparsity.}
We sample context views from a compact trajectory span and select target
views using an occlusion-aware symmetric co-visibility criterion.
The input--target overlap is controlled within \textbf{0.2--0.6}, with
lower overlap corresponding to more challenging extrapolation.
We further evaluate \textbf{1/4/8/16/32/64 input views} under a fixed
context span to isolate the effect of observation density.
More details are provided in the appendix.

\begin{figure*}[t] 
    \centering
    \includegraphics[width=\textwidth]{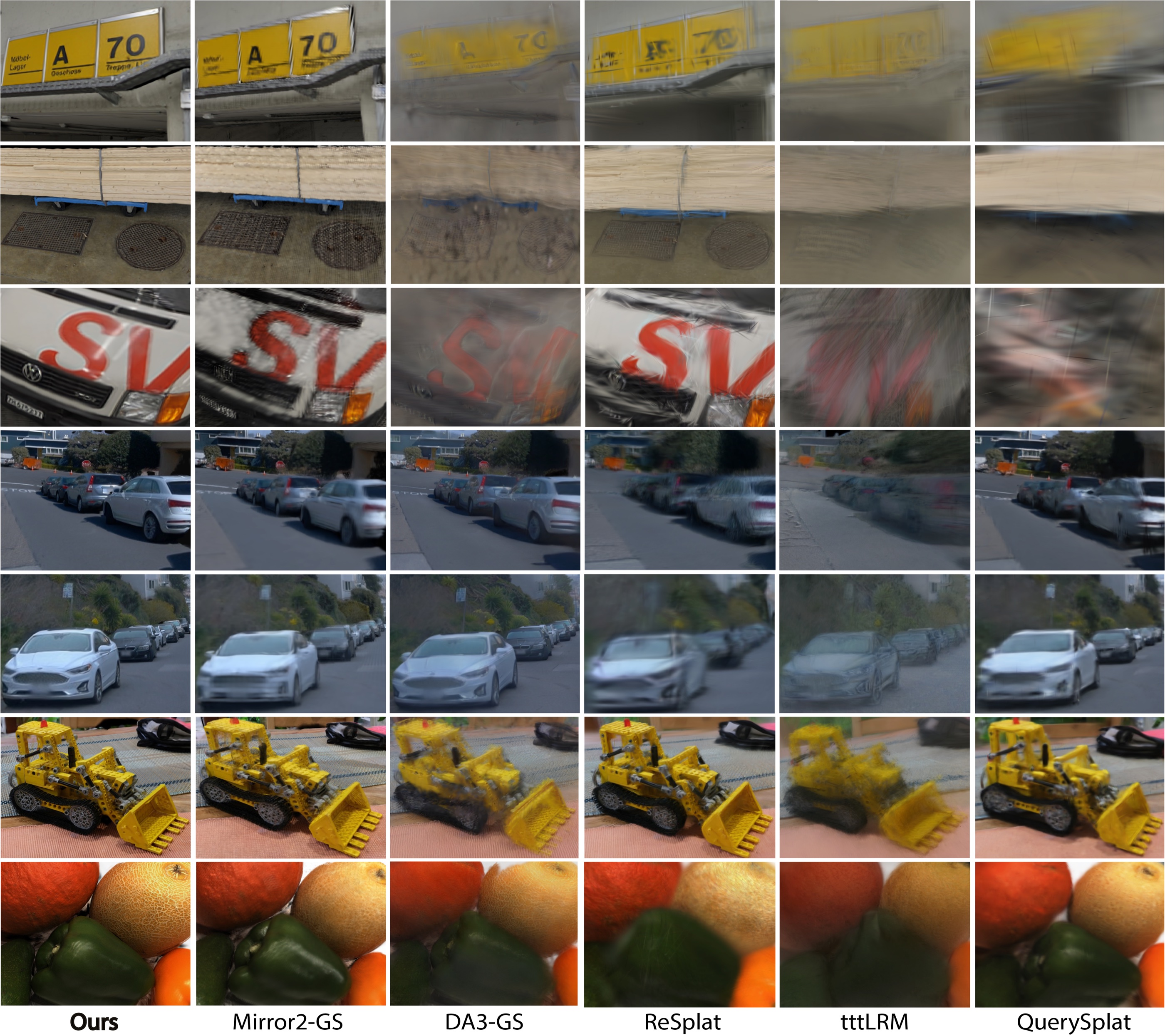}
    \vspace{-0.2in}
    \caption{Qualitative comparisons across diverse scenes and viewpoints with multi-view input.}
    \label{fig:qualitative}
    \vspace{-0.2in}
\end{figure*}

\textbf{Multi-resolution.}
Our benchmark covers a broad resolution range, from approximately \textbf{480p to 1K}, enabling evaluation under both low- and high-resolution settings.

\textbf{Compared methods.}
We evaluate two categories methods according to whether camera poses are provided. In the \emph{w/o Pose} setting, methods reconstruct the scene directly from RGB observations, including AnySplat~\citep{jiang2025anysplat}, YoNoSplat~\citep{ye2026yonosplat}, DA3-GS~\citep{lin2025depth}, Mirror2-GS~\citep{team2026hy}, and QuerySplat~\citep{li2026querysplat}.

\begin{wraptable}{r}{0.50\textwidth}
    \vspace{-0.15in}
    \centering
    \caption{Quantitative comparison of feed-forward 3DGS methods using a \textbf{single input view}. The top-3 results are highlighted as \colorbox{rankfirst}{first}, \colorbox{ranksecond}{second}, and \colorbox{rankthird}{third}.}
    \vspace{0.01in}
    \label{tab:main_comparison_monocular}
    \setlength{\tabcolsep}{4pt}
    \renewcommand{\arraystretch}{1.15}
    \small
    \makebox[\linewidth][c]{%
    \resizebox{\linewidth}{!}{%
    \begin{tabular}{lcccccccccccccc}
        \toprule
        \multirow{2}{*}{\textbf{Method}}
        & \multicolumn{2}{c}{\textbf{ScanNet}}
        & \multicolumn{2}{c}{\textbf{NRGBD}}
        & \multicolumn{2}{c}{\textbf{DL3DV}}
        & \multicolumn{2}{c}{\textbf{Tanks \& Temples}}
        & \multicolumn{2}{c}{\textbf{Mip-NeRF 360}}
        & \multicolumn{2}{c}{\textbf{ETH3D}}
        & \multicolumn{2}{c}{\textbf{DTU}} \\
        \cmidrule(lr){2-3}
        \cmidrule(lr){4-5}
        \cmidrule(lr){6-7}
        \cmidrule(lr){8-9}
        \cmidrule(lr){10-11}
        \cmidrule(lr){12-13}
        \cmidrule(lr){14-15}
        & DISTS$\downarrow$ & LPIPS$\downarrow$
        & DISTS$\downarrow$ & LPIPS$\downarrow$
        & DISTS$\downarrow$ & LPIPS$\downarrow$
        & DISTS$\downarrow$ & LPIPS$\downarrow$
        & DISTS$\downarrow$ & LPIPS$\downarrow$
        & DISTS$\downarrow$ & LPIPS$\downarrow$
        & DISTS$\downarrow$ & LPIPS$\downarrow$ \\
        \midrule
        SHARP
        & \first{0.223} & \first{0.244}
        & \first{0.238} & \first{0.174}
        & \second{0.335} & \second{0.280}
        & \second{0.374} & \second{0.470}
        & \first{0.328} & \first{0.505}
        & \second{0.200} & \second{0.133}
        & \second{0.433} & \second{0.724} \\
        InfiniSplat
        & \third{0.266} & \third{0.280}
        & \third{0.336} & \third{0.345}
        & \third{0.527} & \third{0.596}
        & \third{0.452} & \third{0.531}
        & \third{0.505} & \third{0.654}
        & \third{0.367} & \third{0.376}
        & \third{0.562} & \third{0.754} \\
        \textbf{Ours}
        & \second{0.238} & \second{0.274}
        & \second{0.265} & \second{0.284}
        & \first{0.279} & \first{0.220}
        & \first{0.342} & \first{0.430}
        & \second{0.410} & \second{0.599}
        & \first{0.195} & \first{0.104}
        & \first{0.327} & \first{0.657} \\
        \bottomrule
    \end{tabular}%
    }%
    }
    \vspace{-0.1in}
    \par\vspace{0.1in}
    \centering
    \includegraphics[width=\linewidth]{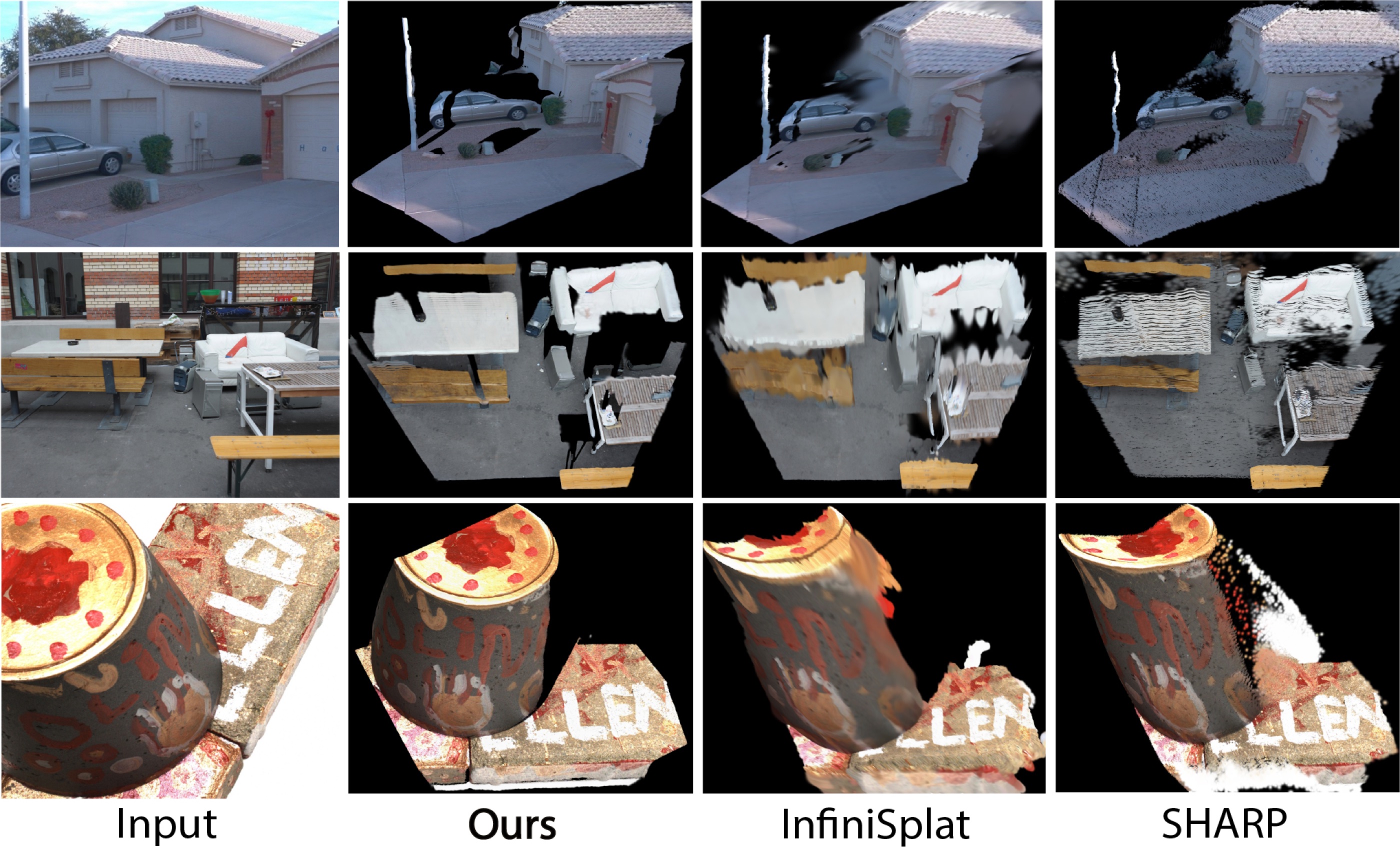}
    \vspace{-0.23in}
    \captionof{figure}{Qualitative comparisons across diverse scenes and viewpoints with monocular input.}
    \label{fig:qualitative_mono}
    \vspace{-0.5in}
\end{wraptable}

In the \emph{w/ Pose} setting, camera poses are additionally provided for model input. We compare against DepthSplat~\citep{xu2025depthsplat}, TokenGS~\citep{ren2026tokengs}, tttLRM~\citep{wang2026tttlrm}, ReSplat~\citep{xu2025resplat}, as well as pose-conditioned variants of DA3-GS~\citep{lin2025depth} and Mirror2-GS~\citep{team2026hy}.
Additionally, we compare against SHARP~\citep{mescheder2026sharp} and InfiniSplat~\citep{wang2026infinisplat}, which target for monocular view synthesis.

\subsection{Quantitative Comparisons}

We report the average metrics over \textbf{4/8/16/32/64 input views} in Table~\ref{tab:main_comparison_4_64view}, specific results under different input-view settings are provided in the appendix. Following SHARP~\citep{mescheder2026sharp}, we primarily report \textit{LPIPS} and \textit{DISTS} in the main paper, as they better reflect rendering quality and higher-frequency details. We also report the results in the monocular setting in Table~\ref{tab:main_comparison_monocular}.
Overall, SplatAnything achieves the strongest performance across most datasets and evaluation settings, demonstrating its superior generalization capabilities.

\subsection{Qualitative Comparisons}

Figs.~\ref{fig:qualitative} and~\ref{fig:qualitative_mono} present qualitative comparisons across diverse scenes and extrapolated viewpoints under multi-view and monocular input settings, covering object-centric captures, indoor environments, and large-scale outdoor scenes. \textbf{SplatAnything} demonstrates a significant advantage in rendering quality: it generalizes robustly across a wide range of scene types and spatial scales, maintains geometric consistency under challenging large viewpoint
extrapolation, and produces sharp, detailed renderings that faithfully recover fine local structures. More results are provided in the appendix.
\subsection{Efficiency Analysis}
We evaluate the inference efficiency of \textbf{SplatAnything} and other methods under different input-view settings. Detailed timing results are provided in the appendix.

\subsection{Ablation Studies}
\begin{figure*}[t] 
    \centering
    \vspace{-0.1in}
    \includegraphics[width=\linewidth]{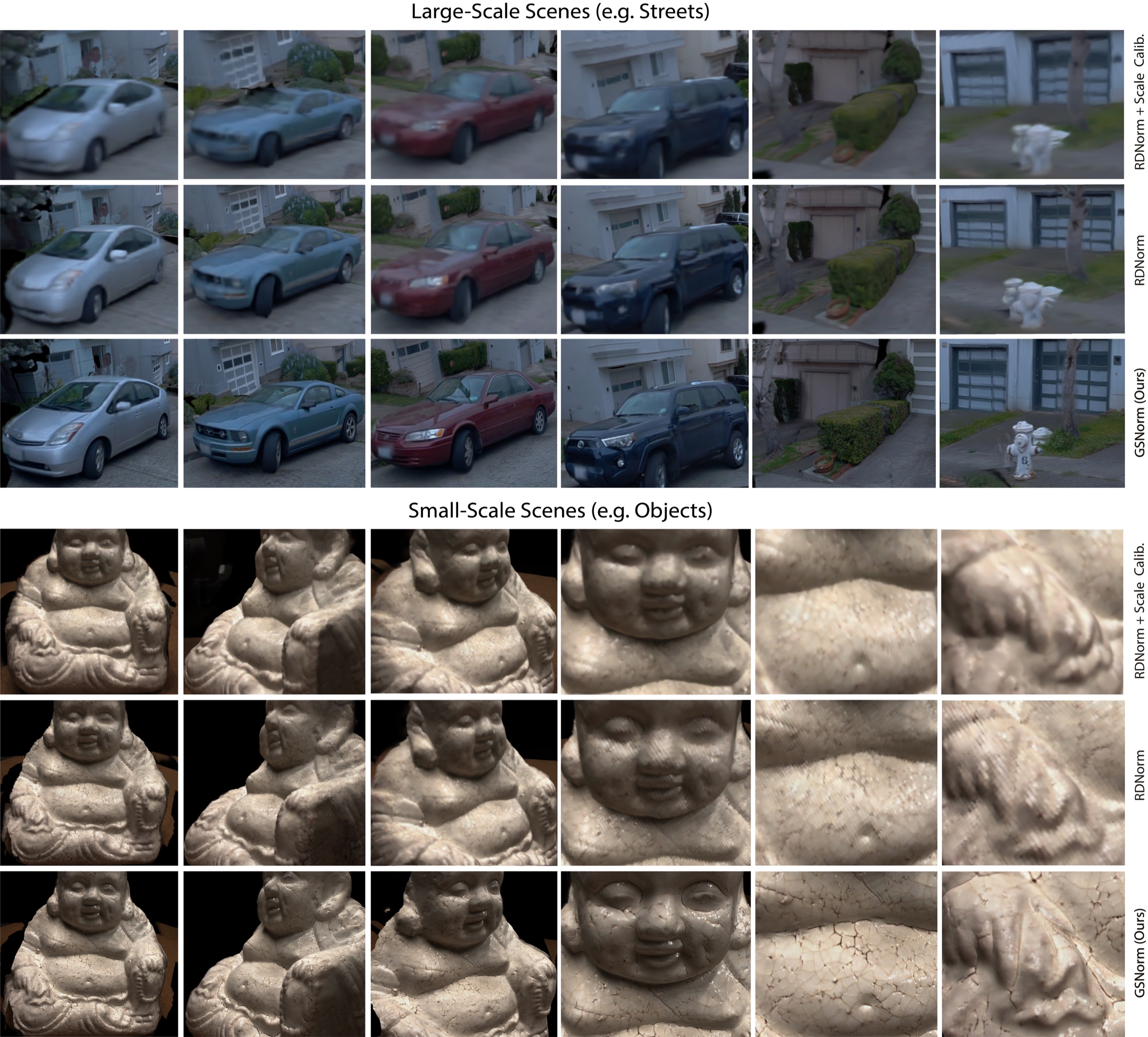}
    \vspace{-0.25in}
    \caption{Ablation studies on the effect of GSNorm across different scene types.}
    \label{fig:ablation}
    \vspace{-0.12in}
\end{figure*}

\paragraph{Effect of Gaussian spacing normalization.}
We ablate the normalization space for Gaussian prediction in
Table~\ref{tab:ablation_normalization_space}, comparing RDNorm,
RDNorm with scale calibration, and GSNorm under the same architecture
and training settings.
For RDNorm with scale calibration, Gaussian prediction remains in the
RDNorm space, while the predicted Gaussian scales are uniformly rescaled
to match the scene's median Gaussian spacing. Both RDNorm variants remain consistently inferior to GSNorm.
Qualitatively, RDNorm tends to produce holes in small object-centric
scenes and over-smoothed results in large outdoor scenes
(Fig.~\ref{fig:ablation}).
Scale calibration alleviates the holes on object-centric scenes, but
often turns them into over-smoothed reconstructions, while outdoor scenes
remain blurry.
These results indicate that post-prediction scale correction cannot
resolve the ambiguity in scale supervision introduced by RDNorm.
By canonicalizing the representative Gaussian spacing scale before Gaussian
prediction, GSNorm provides a better-aligned prediction space and yields
more complete and sharper reconstructions across diverse scenes. 
Additional results are provided in the appendix.

\paragraph{Training design choices}
We also analyze several architecture and training design choices, including
fine-tuning the pretrained geometry backbone, fixing Gaussian centers,
scaling the amount of training data, and using mixed sampling strategies
for Gaussian supervision. Please refer to the appendix for more analysis and results.

\vspace{-0.04in}
\begin{table*}[t]
    \centering
    \caption{Normalization-space ablation across different numbers of input views.}
    \vspace{0.02in}
    \label{tab:ablation_normalization_space}
    \setlength{\tabcolsep}{4pt}
    \renewcommand{\arraystretch}{1.05}
    \small
\makebox[\textwidth][c]{%
\resizebox{\textwidth}{!}{%
    \begin{tabular}{lcccccccccccc}
        \toprule
        \multirow{2}{*}{\textbf{Norm. Space}}
        & \multicolumn{2}{c}{\textbf{Num. Views=1}}
        & \multicolumn{2}{c}{\textbf{Num. Views=4}}
        & \multicolumn{2}{c}{\textbf{Num. Views=8}}
        & \multicolumn{2}{c}{\textbf{Num. Views=16}}
        & \multicolumn{2}{c}{\textbf{Num. Views=32}}
        & \multicolumn{2}{c}{\textbf{Num. Views=64}} \\
        \cmidrule(lr){2-3}
        \cmidrule(lr){4-5}
        \cmidrule(lr){6-7}
        \cmidrule(lr){8-9}
        \cmidrule(lr){10-11}
        \cmidrule(lr){12-13}
        & DISTS$\downarrow$ & LPIPS$\downarrow$
        & DISTS$\downarrow$ & LPIPS$\downarrow$
        & DISTS$\downarrow$ & LPIPS$\downarrow$
        & DISTS$\downarrow$ & LPIPS$\downarrow$
        & DISTS$\downarrow$ & LPIPS$\downarrow$
        & DISTS$\downarrow$ & LPIPS$\downarrow$ \\
        \midrule
        Radial Distance Norm.
        & 0.308 & 0.394
        & 0.223 & 0.278
        & 0.223 & 0.272
        & 0.218 & 0.268
        & 0.225 & 0.277
        & 0.224 & 0.250 \\
        Radial Distance Norm + Scale Calibration.
        & 0.323 & 0.400
        & 0.249 & 0.309
        & 0.252 & 0.308
        & 0.245 & 0.300
        & 0.253 & 0.305
        & 0.234 & 0.265 \\
        \textbf{Gaussian Spacing Norm. (Ours)}
        & \textbf{0.294} & \textbf{0.367}
        & \textbf{0.198} & \textbf{0.245}
        & \textbf{0.196} & \textbf{0.237}
        & \textbf{0.190} & \textbf{0.232}
        & \textbf{0.197} & \textbf{0.241}
        & \textbf{0.217} & \textbf{0.229} \\
        \bottomrule
    \end{tabular}%
}%
}
\end{table*}


\section{Conclusion}

\label{sec:conclusion}
We identified a key limitation of existing methods: \emph{Radial Distance
Normalization (RDNorm)} does not directly normalize Gaussian spacing,
leading to ambiguity in Gaussian-scale supervision across scene scales and
camera configurations. To address this issue, we introduced
\textbf{SplatAnything} with \textit{Gaussian Spacing Normalization (GSNorm)},
which derives a scene-level normalization factor from a representative
statistic of Gaussian spacing and provides a more consistent spatial
reference for Gaussian prediction. We further established a systematic
cross-dataset benchmark covering different input sparsity levels and
challenging extrapolated viewpoints. Experiments demonstrate that
SplatAnything achieves the strongest performance across most evaluation
settings, highlighting the effectiveness of GSNorm and marking a
substantial step toward feed-forward 3DGS.

\paragraph{Limitations and Future Work.}
SplatAnything relies on geometric priors from a geometry foundation model
and may degrade when reliable geometry cannot be recovered from extremely
sparse or challenging inputs. It also primarily reconstructs regions
visible in the input views, which can leave holes in unobserved areas.
Future work will explore more robust geometry estimation and generative
priors for completing unobserved regions.

\subsection*{AI use statement}

In this work, we used generative AI tools to assist with writing, language
polishing, and code development. In particular, generative AI was used to
improve the clarity, grammar, and presentation of the manuscript, and to
assist with parts of code implementation and debugging under the authors'
direction. All AI-assisted text was carefully reviewed and revised by the
authors, and all AI-assisted code was manually inspected, tested, and
verified for correctness.
Generative AI was not used to independently originate the research ideas,
formulate the research questions, or design the core technical
contributions of this work. We also did not use generative AI tools to
generate experimental or training data, or to independently search for,
select, or determine the references used in the paper. The authors take
full responsibility for the final content of this work, including all
text, technical claims, experimental results, and code produced with the
assistance of generative AI.

\nocite{*}

\bibliography{iclr2027_conference}
\bibliographystyle{iclr2027_conference}

\clearpage
\appendix
\section{Appendix}
\label{sec:appendix}

\subsection{Training Data.}

The training data used for \textbf{SplatAnything} is listed in Table~\ref{tab:training_data}.

\vspace{-0.15in}
\begin{table*}[h]
    \centering
    \caption{
        Details of the training datasets used by SplatAnything.
    }
    \vspace{0.1in}
    \label{tab:training_data}
    \setlength{\tabcolsep}{9pt}
    \renewcommand{\arraystretch}{1.0}
    \begin{tabular}{@{}llcc@{}}
        \toprule
        \textbf{Data Type} & \textbf{Dataset} & \textbf{\# Scenes} & \textbf{Resolution ($H \times W$)} \\
        \midrule
        \multirow{7}{*}{Synthetic}
        & HyperSim~\citep{roberts2021hypersim}     & 364 & 768$\times$1024 \\
        & UnrealStereo4K~\citep{tosi2021smd}     & 18 & 1080$\times$1920 \\
        & OmniObject3D~\citep{wu2023omniobject3d} & 5914 & 800$\times$800 \\
        & Rawmantic3D~\citep{rawmantic2026data}    & 30 & 1080$\times$1920 \\
        & OmniWorld~\citep{zhou2025omniworld}     & 10464 & 720$\times$1280 \\
        & GTA-SfM~\citep{wang2020flow}      & 200 & 480$\times$640 \\
        & MatrixCity~\citep{li2023matrixcity}   & 8 & 1080$\times$1920 \\
        & TartanGround~\citep{patel2025tartanground} & 635 & 640$\times$640 \\
        \midrule
        \multirow{7}{*}{Real-world}
        & ScanNet++~\citep{yeshwanth2023scannet++}   & 936 & 696$\times$1035 \\
        & Matterport3D~\citep{chang2017matterport3d}  & 90 & 1024$\times$1280 \\
        & ARKitScenes~\citep{baruch2021arkitscenes} & 2501 & 480$\times$640 \\
        & Taskonomy~\citep{zamir2018taskonomy}   & 537 & 512$\times$512 \\
        & MapFree~\citep{arnold2022map}     & 460 & 720$\times$540 \\
        & BlendedMVS~\citep{yao2020blendedmvs}  & 502 & 384$\times$504 \\
        \bottomrule
    \end{tabular}
\end{table*}
\vspace{-0.15in}

\subsection{Data Construction Strategies.}
\begin{figure*}[t] 
    \centering
    \includegraphics[width=\textwidth]{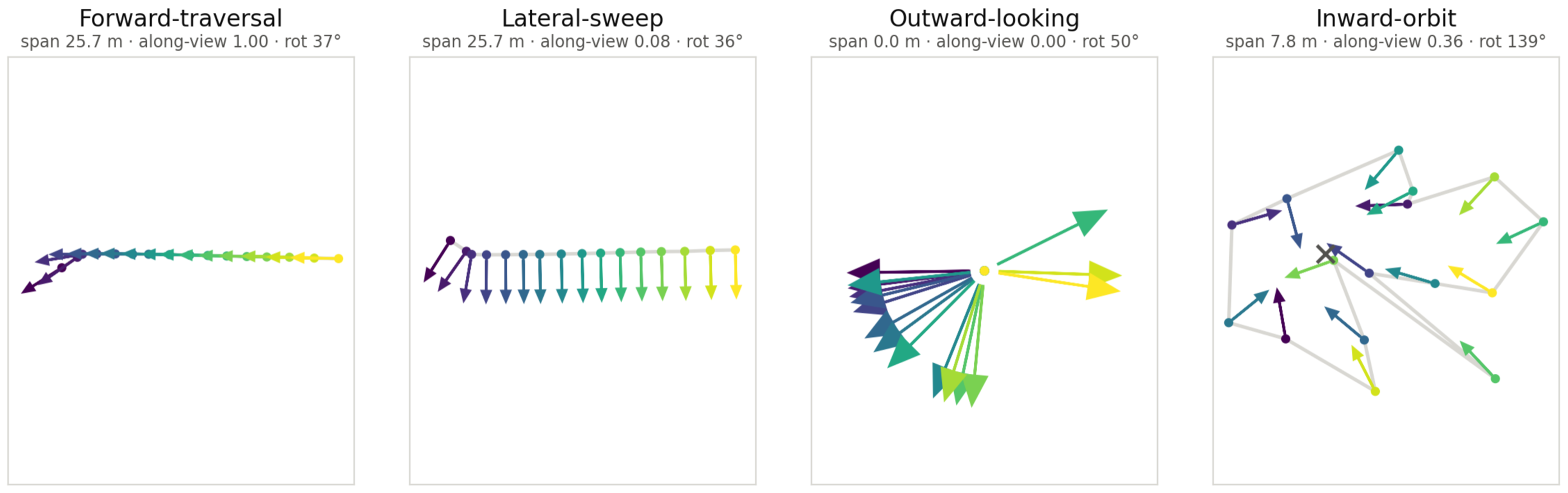}
    \vspace{-0.25in}
    \caption{
        \textbf{Source-view trajectory patterns} ($16$ views per panel), visualized in an approximately top-down projection.
        Dots and arrows denote camera centers and viewing directions, respectively;
        the gray polyline connects consecutive views, and color indicates capture order.
        For \emph{inward-orbit}, $\times$ marks the approximate viewing center.
    }
    \label{fig:source_view}
\end{figure*}

We design multiple strategies to construct the training pairs for \textbf{SplatAnything}. 
\paragraph{Source-View Sampling Strategy.}
We construct source views to cover diverse camera-motion patterns while maintaining sufficient geometric overlap for reliable reconstruction by the geometry foundation model. We consider four representative trajectory
patterns: \emph{forward-traversal}, where the camera moves primarily along the viewing direction; \emph{lateral-sweep}, where the camera translates mainly sideways while maintaining a similar viewing direction; \emph{outward-looking}, where the camera remains approximately stationary and rotates to observe different directions; and \emph{inward-orbit}, where the camera moves around a local region while maintaining an inward-facing view toward that region. For trajectories with substantial translation, we further vary the sampling stride to produce either densely overlapping or wide-baseline source views, covering different levels of parallax and input sparsity. During training, the trajectory pattern and sampling stride are randomly selected from a dataset-specific mixture, with distance and stride budgets adapted to each dataset. Validation uses a fixed deterministic sampling configuration for reproducibility. Examples of the different source-view sampling patterns are shown in Fig.~\ref{fig:source_view}.

\begin{figure*}[t] 
    \centering
    \includegraphics[width=\textwidth]{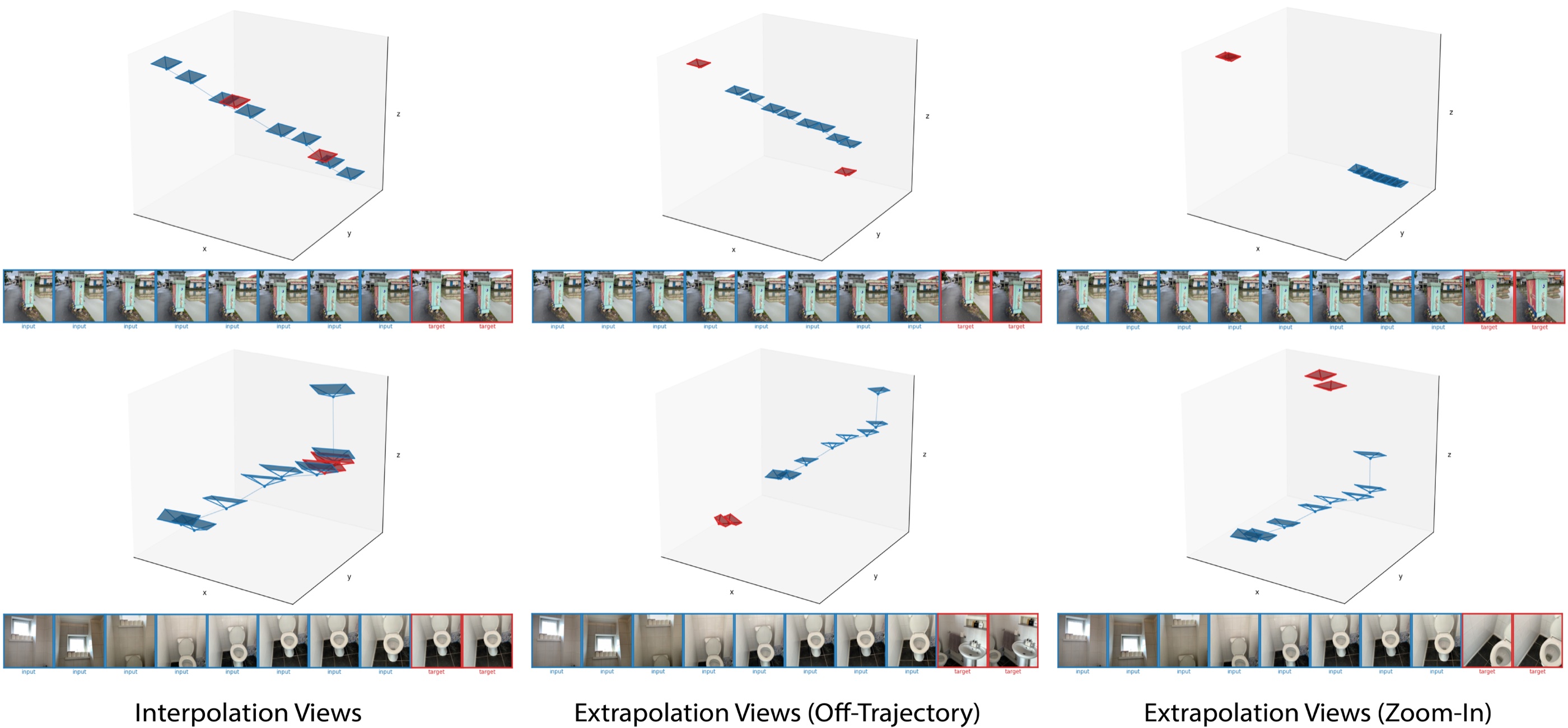}
    \vspace{-0.25in}
    \caption{
        \textbf{Target-view sampling modes} used during training.
        Blue cameras denote the source views provided as input, while red cameras denote the target views used for rendering supervision.
    }
    \label{fig:target_view}
    \vspace{-0.15in}
\end{figure*}

\paragraph{Sampling strategy for target views.}
Our target-view sampling aims to provide supervision across diverse viewpoint changes, ranging from interpolation to challenging extrapolation, in order to improve the viewpoint generalization ability of our model. For \emph{interpolation views}, we select a held-out view between the input views; for \emph{extrapolation views}, we randomly sample diverse extrapolation patterns, including advancing along the camera trajectory to inspect local scene details at a closer range, as well as moving away from the observed trajectory to view the scene from novel positions and orientations. The visualization of different target-view sampling modes are shown in Fig.~\ref{fig:target_view}.

\subsection{Evaluation Benchmark Details}
\label{sec:appendix_evaluation_benchmark}

We provide additional details of the evaluation benchmark introduced in
the main paper. Our benchmark spans seven datasets and evaluates different
input-view counts of $4$, $8$, $16$, $32$, and $64$. The protocol is
designed to independently control input-view sparsity and target-view
difficulty, while ensuring identical evaluation conditions across methods.

\paragraph{Source-view construction.}
Instead of distributing the source views uniformly over the entire
capture, we sample them from a compact contiguous arc of the camera
trajectory. This design leaves a meaningful portion of the capture
outside the observed region, making it possible to evaluate genuinely
extrapolated novel views rather than views that are always bracketed by
the inputs.
To make different input-view counts directly comparable, we adopt a
\emph{fixed-span} strategy. For each scene, we first determine the
trajectory span at the sparsest setting of $4$ source views. We select
the widest arc whose adjacent source-view pairs maintain sufficient
geometric overlap, measured by symmetric co-visibility; specifically,
the $20$-th percentile of adjacent-pair co-visibility is required to be
at least $0.35$. The resulting span and offset are then reused for
$8$, $16$, $32$, and $64$ views, with additional views inserted within
the same trajectory range. Therefore, all view-count settings share the
same spatial coverage, and only the observation density changes. This
allows the effect of input sparsity to be evaluated without conflating
it with changes in scene coverage.

\paragraph{Target-view selection.}
Target views are sampled from frames outside the source-view arc and are
selected according to an occlusion-aware symmetric co-visibility measure.
For each candidate target view, valid pixels are back-projected using
ground-truth depth and projected into the source views. Pixels that are
geometrically consistent with at least one source view are treated as
observed, while occluded and out-of-frustum pixels reduce the overlap
score; pixels for which visibility cannot be determined due to invalid
depth are excluded from the statistic.

We compute co-visibility in both projection directions and define
\begin{equation}
    C_{\mathrm{sym}}
    =
    \min
    \left(
        C_{\mathrm{fwd}},
        C_{\mathrm{bwd}}
    \right).
\end{equation}
The symmetric formulation is important for asymmetric camera motions,
such as moving toward the scene, where a one-sided overlap measure can
remain high even under a substantial viewpoint change. We evaluate up
to $40$ candidate target views per scene and retain those whose symmetric
co-visibility lies in
\begin{equation}
    0.20 \leq C_{\mathrm{sym}} < 0.60.
\end{equation}
Up to four target views are selected from this interval, prioritizing
lower co-visibility and, when necessary, larger parallax. In this way,
target-view difficulty is explicitly controlled by geometric overlap
rather than by temporal or camera-index distance alone, providing a
consistent extrapolation regime across datasets and methods.

\paragraph{Evaluation without input poses.}
For pose-free methods, we estimate a scene-level similarity transformation between the predicted source-camera poses and their ground-truth counterparts, and apply this transformation to the predicted Gaussian representation. The aligned Gaussians are then rendered from the ground-truth target-camera poses, and the resulting images are evaluated against the corresponding target views. Ground-truth source poses are used only for this post-hoc
alignment and are not provided to the model during inference. This
allows pose-free and pose-conditioned methods to be evaluated on
identical target viewpoints in the same ground-truth coordinate system.


\subsection{Implementation Details}

\paragraph{GS Head Architecture.}
The architecture of the GS Head follows WorldMirror~2.0~\citep{team2026hy}.
For Gaussian attribute prediction, we parameterize the Gaussian scale in
log space and apply an exponential activation, followed by clamping to a
maximum value of $0.3$:
\begin{equation}
    \mathbf{s}
    =
    \min\left(\exp(\hat{\mathbf{s}}),\, 0.3\right).
\end{equation}
Gaussian rotations are represented as quaternions and normalized to unit
length,
\begin{equation}
    \mathbf{q}
    =
    \frac{\hat{\mathbf{q}}}
    {\|\hat{\mathbf{q}}\|_2 + \epsilon}.
\end{equation}
Opacity is activated using the sigmoid function, constraining them to $[0,1]$. The spherical harmonic coefficients are used directly without an additional nonlinear activation.

\paragraph{Coordinate alignment.}
During training, the predicted Gaussians are rendered using the
ground-truth target camera poses for supervision. However, the camera
trajectory and depth predicted by the geometry model generally have an
arbitrary global scale relative to the ground-truth trajectory. Directly
combining the predicted depth with the ground-truth poses would therefore
lead to inconsistent geometry and camera scales, resulting in misaligned
rendering supervision.
To resolve this scale ambiguity, we estimate a scene-level alignment factor
from the predicted and ground-truth source-camera trajectories. Specifically,
we compute the ratio between corresponding pairwise camera baselines and use
their median as a robust global scale factor. The predicted depth maps are
then rescaled by this factor and back-projected using the ground-truth
source-camera poses to construct Gaussian centers in the ground-truth
coordinate system. The resulting Gaussians can therefore be rendered
directly using the ground-truth target-camera poses for supervision. This
alignment is applied to both the pose-free and pose-conditioned variants,
since the predicted camera trajectory can exhibit a global scale discrepancy
with respect to the ground truth in either setting.

\paragraph{Voxel Merge Strategy.}
GSNorm canonicalizes the representative single-view point spacing across scenes and camera configurations, but it does not by itself eliminate the variation in primitive density caused by multi-view accumulation. Since each valid pixel predicts a Gaussian primitive, increasing the number of input views over the same spatial region can introduce multiple Gaussians at nearby 3D locations, making the resulting primitive density dependent on the number of overlapping observations. To reduce such redundancy, we apply a spatial voxel merging strategy following AnySplat~\citep{jiang2025anysplat}. Specifically, we predict a per-pixel weight map $w^{p}=\{w_i^{p}\}_{i=1}^{N}$ and merge redundant per-view Gaussians within local voxels. This produces the final unified 3D Gaussian scene representation $\mathcal{G}$.
GSNorm and voxel merging therefore address two complementary sources of scale variation: GSNorm provides a consistent characteristic sampling scale for Gaussian prediction across scenes, whereas voxel merging regularizes the accumulation of primitives caused by overlapping multi-view observations. Together, they allow the model to operate robustly over varying numbers of input views.

\begin{table*}[t]
\centering
\caption{Quantitative comparison across seven benchmark datasets using \textbf{4 input views}. The top-3 results are highlighted as \colorbox{rankfirst}{first}, \colorbox{ranksecond}{second}, and \colorbox{rankthird}{third}.}

\label{tab:main_comparison_4view}
\setlength{\tabcolsep}{1pt}
\renewcommand{\arraystretch}{1.2}
\small
    \makebox[\textwidth][c]{%
    \resizebox{\textwidth}{!}{%
    \begin{tabular}{clcccccccccccccc}
        \toprule
        \multirow{2}{*}{\textbf{Input}}
        & \multirow{2}{*}{\textbf{Method}}
        & \multicolumn{2}{c}{\textbf{ScanNet}}
        & \multicolumn{2}{c}{\textbf{NRGBD}}
        & \multicolumn{2}{c}{\textbf{DL3DV}}
        & \multicolumn{2}{c}{\textbf{Tanks \& Temples}}
        & \multicolumn{2}{c}{\textbf{Mip-NeRF 360}}
        & \multicolumn{2}{c}{\textbf{ETH3D}}
        & \multicolumn{2}{c}{\textbf{DTU}} \\
        \cmidrule(lr){3-4}
        \cmidrule(lr){5-6}
        \cmidrule(lr){7-8}
        \cmidrule(lr){9-10}
        \cmidrule(lr){11-12}
        \cmidrule(lr){13-14}
        \cmidrule(lr){15-16}
        & & DISTS$\downarrow$ & LPIPS$\downarrow$
        & DISTS$\downarrow$ & LPIPS$\downarrow$
        & DISTS$\downarrow$ & LPIPS$\downarrow$
        & DISTS$\downarrow$ & LPIPS$\downarrow$
        & DISTS$\downarrow$ & LPIPS$\downarrow$
        & DISTS$\downarrow$ & LPIPS$\downarrow$
        & DISTS$\downarrow$ & LPIPS$\downarrow$ \\
        \midrule
        \multirow{6}{*}{\rotatebox[origin=c]{90}{w/o Pose}}
        & AnySplat
        & 0.289 & 0.467
        & 0.249 & 0.377
        & 0.350 & 0.319
        & 0.624 & 0.777
        & 0.575 & 0.825
        & 0.296 & 0.173
        & 0.491 & 0.819 \\
        & YoNoSplat
        & 0.253 & \third{0.231}
        & 0.272 & \third{0.216}
        & 0.366 & \third{0.213}
        & \third{0.427} & \second{0.340}
        & 0.398 & \third{0.573}
        & 0.325 & \second{0.036}
        & 0.484 & 0.753 \\
        & DA3-GS
        & 0.291 & 0.322
        & \third{0.243} & 0.219
        & \third{0.320} & 0.222
        & 0.441 & 0.417
        & \third{0.388} & 0.582
        & \third{0.249} & 0.090
        & \third{0.386} & \third{0.662} \\
        & Mirror2-GS
        & \second{0.188} & \second{0.218}
        & \second{0.234} & 0.229
        & \second{0.238} & \second{0.177}
        & \second{0.292} & \third{0.348}
        & \second{0.263} & \second{0.525}
        & \second{0.214} & 0.080
        & \second{0.328} & \second{0.645} \\
        & QuerySplat
        & \third{0.241} & 0.237
        & 0.258 & \second{0.206}
        & 0.349 & 0.285
        & 0.438 & 0.396
        & 0.427 & 0.652
        & 0.296 & \first{0.031}
        & 0.438 & 0.716 \\
        & \textbf{Ours}
        & \first{0.185} & \first{0.212}
        & \first{0.199} & \first{0.185}
        & \first{0.221} & \first{0.150}
        & \first{0.240} & \first{0.301}
        & \first{0.226} & \first{0.472}
        & \first{0.151} & \third{0.047}
        & \first{0.229} & \first{0.526} \\
        \midrule
        \multirow{7}{*}{\rotatebox[origin=c]{90}{w/ Pose}}
        & DepthSplat
        & 0.406 & 0.326
        & 0.393 & 0.278
        & 0.369 & 0.188
        & 0.521 & 0.378
        & 0.642 & 0.695
        & 0.523 & 0.084
        & 0.402 & 0.682 \\
        & TokenGS
        & 0.408 & 0.287
        & 0.352 & 0.279
        & 0.375 & 0.202
        & 0.534 & 0.354
        & 0.516 & 0.708
        & 0.516 & \third{0.049}
        & 0.651 & 0.871 \\
        & tttLRM
        & 0.354 & 0.378
        & 0.402 & 0.421
        & 0.351 & 0.223
        & \third{0.406} & \third{0.336}
        & 0.404 & 0.590
        & 0.250 & 0.066
        & 0.463 & 0.713 \\
        & ReSplat
        & \third{0.256} & \third{0.224}
        & \third{0.249} & \second{0.184}
        & \third{0.268} & \first{0.140}
        & 0.447 & \second{0.296}
        & 0.402 & \third{0.578}
        & \third{0.219} & \first{0.028}
        & 0.522 & 0.773 \\
        & DA3-GS (Pose Cond.)
        & 0.285 & 0.314
        & 0.253 & 0.237
        & 0.306 & 0.217
        & 0.447 & 0.402
        & \third{0.381} & 0.585
        & 0.245 & 0.093
        & \third{0.380} & \third{0.677} \\
        & Mirror2-GS (Pose Cond.)
        & \second{0.188} & \second{0.219}
        & \second{0.227} & \third{0.209}
        & \second{0.232} & \third{0.170}
        & \second{0.289} & 0.344
        & \second{0.257} & \second{0.504}
        & \second{0.213} & 0.069
        & \second{0.321} & \second{0.587} \\
        & \textbf{Ours (Pose Cond.)}
        & \first{0.185} & \first{0.210}
        & \first{0.189} & \first{0.161}
        & \first{0.216} & \second{0.142}
        & \first{0.238} & \first{0.291}
        & \first{0.217} & \first{0.448}
        & \first{0.134} & \second{0.031}
        & \first{0.209} & \first{0.432} \\
        \bottomrule
    \end{tabular}%
    }%
    }
\vspace{-0.15in}
    \par\vspace{3pt}
\centering
\caption{Quantitative comparison across seven benchmark datasets using \textbf{8 input views}. The top-3 results are highlighted as \colorbox{rankfirst}{first}, \colorbox{ranksecond}{second}, and \colorbox{rankthird}{third}.}

\label{tab:main_comparison_8view}
\setlength{\tabcolsep}{1pt}
\renewcommand{\arraystretch}{1.2}
\small
\makebox[\textwidth][c]{%
\resizebox{\textwidth}{!}{%
\begin{tabular}{clcccccccccccccc}
    \toprule
    \multirow{2}{*}{\textbf{Input}}
    & \multirow{2}{*}{\textbf{Method}}
    & \multicolumn{2}{c}{\textbf{ScanNet}}
    & \multicolumn{2}{c}{\textbf{NRGBD}}
    & \multicolumn{2}{c}{\textbf{DL3DV}}
    & \multicolumn{2}{c}{\textbf{Tanks \& Temples}}
    & \multicolumn{2}{c}{\textbf{Mip-NeRF 360}}
    & \multicolumn{2}{c}{\textbf{ETH3D}}
    & \multicolumn{2}{c}{\textbf{DTU}} \\
    \cmidrule(lr){3-4}
    \cmidrule(lr){5-6}
    \cmidrule(lr){7-8}
    \cmidrule(lr){9-10}
    \cmidrule(lr){11-12}
    \cmidrule(lr){13-14}
    \cmidrule(lr){15-16}
    & & DISTS$\downarrow$ & LPIPS$\downarrow$
    & DISTS$\downarrow$ & LPIPS$\downarrow$
    & DISTS$\downarrow$ & LPIPS$\downarrow$
    & DISTS$\downarrow$ & LPIPS$\downarrow$
    & DISTS$\downarrow$ & LPIPS$\downarrow$
    & DISTS$\downarrow$ & LPIPS$\downarrow$
    & DISTS$\downarrow$ & LPIPS$\downarrow$ \\
    \midrule
    \multirow{6}{*}{\rotatebox[origin=c]{90}{w/o Pose}}
    & AnySplat
    & 0.291 & 0.447
    & 0.238 & 0.356
    & 0.364 & 0.322
    & 0.635 & 0.801
    & 0.538 & 0.806
    & 0.287 & 0.143
    & 0.476 & 0.808 \\
    & YoNoSplat
    & 0.262 & \third{0.234}
    & 0.280 & 0.226
    & 0.374 & \third{0.219}
    & \third{0.423} & \second{0.343}
    & 0.406 & 0.592
    & 0.330 & \second{0.027}
    & 0.453 & 0.727 \\
    & DA3-GS
    & 0.302 & 0.326
    & \third{0.237} & \third{0.219}
    & \third{0.329} & 0.226
    & 0.442 & 0.409
    & \third{0.393} & \third{0.591}
    & \third{0.241} & 0.075
    & \third{0.365} & \third{0.604} \\
    & Mirror2-GS
    & \second{0.190} & \second{0.225}
    & \second{0.232} & 0.226
    & \second{0.246} & \second{0.191}
    & \second{0.302} & \third{0.356}
    & \second{0.283} & \second{0.543}
    & \second{0.219} & 0.072
    & \second{0.334} & \second{0.600} \\
    & QuerySplat
    & \third{0.236} & 0.241
    & 0.259 & \second{0.211}
    & 0.344 & 0.299
    & 0.431 & 0.398
    & 0.424 & 0.664
    & 0.294 & \first{0.025}
    & 0.444 & 0.710 \\
    & \textbf{Ours}
    & \first{0.185} & \first{0.218}
    & \first{0.197} & \first{0.184}
    & \first{0.221} & \first{0.159}
    & \first{0.246} & \first{0.306}
    & \first{0.235} & \first{0.482}
    & \first{0.156} & \third{0.040}
    & \first{0.216} & \first{0.467} \\
    \midrule
    \multirow{7}{*}{\rotatebox[origin=c]{90}{w/ Pose}}
    & DepthSplat
    & 0.301 & 0.281
    & 0.293 & 0.221
    & 0.308 & \second{0.181}
    & 0.414 & 0.348
    & 0.616 & 0.701
    & 0.509 & 0.066
    & \third{0.366} & 0.632 \\
    & TokenGS
    & 0.387 & 0.277
    & 0.330 & 0.268
    & 0.378 & 0.208
    & 0.528 & 0.355
    & 0.517 & 0.734
    & 0.382 & \third{0.041}
    & 0.606 & 0.863 \\
    & tttLRM
    & 0.270 & 0.283
    & 0.326 & 0.290
    & 0.314 & 0.192
    & \third{0.394} & \third{0.345}
    & \third{0.356} & \third{0.560}
    & 0.246 & 0.071
    & 0.369 & \third{0.611} \\
    & ReSplat
    & \third{0.267} & \third{0.237}
    & 0.253 & \second{0.186}
    & \third{0.288} & \first{0.152}
    & 0.415 & \first{0.290}
    & 0.372 & 0.565
    & \third{0.220} & \first{0.023}
    & 0.466 & 0.715 \\
    & DA3-GS (Pose Cond.)
    & 0.292 & 0.317
    & \third{0.240} & 0.226
    & 0.306 & 0.217
    & 0.459 & 0.386
    & 0.389 & 0.595
    & 0.239 & 0.079
    & 0.377 & 0.632 \\
    & Mirror2-GS (Pose Cond.)
    & \second{0.193} & \second{0.230}
    & \second{0.226} & \third{0.211}
    & \second{0.239} & \third{0.183}
    & \second{0.300} & 0.350
    & \second{0.274} & \second{0.526}
    & \second{0.219} & 0.063
    & \second{0.301} & \second{0.516} \\
    & \textbf{Ours (Pose Cond.)}
    & \first{0.186} & \first{0.217}
    & \first{0.186} & \first{0.162}
    & \first{0.217} & \first{0.152}
    & \first{0.245} & \second{0.294}
    & \first{0.221} & \first{0.457}
    & \first{0.145} & \second{0.030}
    & \first{0.170} & \first{0.349} \\
    \bottomrule
\end{tabular}%
}%
}
\vspace{-0.15in}
    \par\vspace{3pt}
\centering
\caption{Quantitative comparison across seven benchmark datasets using \textbf{16 input views}. The top-3 results are highlighted as \colorbox{rankfirst}{first}, \colorbox{ranksecond}{second}, and \colorbox{rankthird}{third}.}

\label{tab:main_comparison_16view}
\setlength{\tabcolsep}{1pt}
\renewcommand{\arraystretch}{1.2}
\small
    \makebox[\textwidth][c]{%
    \resizebox{\textwidth}{!}{%
    \begin{tabular}{clcccccccccccccc}
        \toprule
        \multirow{2}{*}{\textbf{Input}}
        & \multirow{2}{*}{\textbf{Method}}
        & \multicolumn{2}{c}{\textbf{ScanNet}}
        & \multicolumn{2}{c}{\textbf{NRGBD}}
        & \multicolumn{2}{c}{\textbf{DL3DV}}
        & \multicolumn{2}{c}{\textbf{Tanks \& Temples}}
        & \multicolumn{2}{c}{\textbf{Mip-NeRF 360}}
        & \multicolumn{2}{c}{\textbf{ETH3D}}
        & \multicolumn{2}{c}{\textbf{DTU}} \\
        \cmidrule(lr){3-4}
        \cmidrule(lr){5-6}
        \cmidrule(lr){7-8}
        \cmidrule(lr){9-10}
        \cmidrule(lr){11-12}
        \cmidrule(lr){13-14}
        \cmidrule(lr){15-16}
        & & DISTS$\downarrow$ & LPIPS$\downarrow$
        & DISTS$\downarrow$ & LPIPS$\downarrow$
        & DISTS$\downarrow$ & LPIPS$\downarrow$
        & DISTS$\downarrow$ & LPIPS$\downarrow$
        & DISTS$\downarrow$ & LPIPS$\downarrow$
        & DISTS$\downarrow$ & LPIPS$\downarrow$
        & DISTS$\downarrow$ & LPIPS$\downarrow$ \\
        \midrule
        \multirow{6}{*}{\rotatebox[origin=c]{90}{w/o Pose}}
        & AnySplat
        & 0.294 & 0.438
        & 0.234 & 0.348
        & 0.370 & 0.322
        & 0.656 & 0.811
        & 0.570 & 0.812
        & 0.286 & 0.195
        & 0.534 & 0.806 \\
        & YoNoSplat
        & 0.268 & \third{0.238}
        & 0.287 & 0.242
        & 0.381 & \third{0.222}
        & 0.429 & \second{0.342}
        & 0.400 & \third{0.576}
        & 0.313 & \second{0.029}
        & 0.503 & 0.729 \\
        & DA3-GS
        & 0.298 & 0.328
        & \second{0.227} & \second{0.211}
        & \third{0.341} & 0.228
        & 0.464 & 0.407
        & \third{0.392} & 0.582
        & \third{0.209} & 0.092
        & \third{0.393} & \third{0.601} \\
        & Mirror2-GS
        & \second{0.190} & \second{0.232}
        & \third{0.233} & 0.235
        & \second{0.247} & \second{0.185}
        & \second{0.308} & \third{0.354}
        & \second{0.274} & \second{0.531}
        & \second{0.181} & 0.047
        & \second{0.353} & \second{0.599} \\
        & QuerySplat
        & \third{0.234} & 0.250
        & 0.263 & \third{0.217}
        & 0.344 & 0.297
        & \third{0.428} & \third{0.400}
        & 0.410 & 0.654
        & 0.285 & \first{0.026}
        & 0.409 & 0.651 \\
        & \textbf{Ours}
        & \first{0.180} & \first{0.217}
        & \first{0.191} & \first{0.183}
        & \first{0.219} & \first{0.155}
        & \first{0.252} & \first{0.315}
        & \first{0.227} & \first{0.467}
        & \first{0.121} & \third{0.037}
        & \first{0.238} & \first{0.484} \\
        \midrule
        \multirow{7}{*}{\rotatebox[origin=c]{90}{w/ Pose}}
        & DepthSplat
        & \third{0.248} & 0.264
        & 0.302 & 0.247
        & 0.305 & 0.177
        & \third{0.351} & \third{0.345}
        & 0.528 & 0.634
        & 0.516 & 0.087
        & \third{0.337} & \third{0.519} \\
        & TokenGS
        & 0.381 & 0.281
        & 0.329 & 0.276
        & 0.377 & 0.207
        & 0.535 & 0.354
        & 0.514 & 0.728
        & 0.354 & \third{0.035}
        & 0.562 & 0.805 \\
        & tttLRM
        & 0.274 & 0.282
        & 0.304 & 0.260
        & 0.310 & 0.184
        & 0.388 & 0.347
        & \third{0.336} & \third{0.529}
        & 0.205 & 0.048
        & 0.346 & 0.544 \\
        & ReSplat
        & 0.277 & \third{0.246}
        & 0.266 & \second{0.207}
        & 0.307 & \second{0.161}
        & 0.398 & \first{0.282}
        & 0.355 & 0.537
        & \third{0.196} & \first{0.022}
        & 0.434 & 0.657 \\
        & DA3-GS (Pose Cond.)
        & 0.288 & 0.318
        & \second{0.229} & \third{0.219}
        & \third{0.299} & 0.212
        & 0.480 & 0.384
        & 0.389 & 0.584
        & 0.207 & 0.094
        & 0.391 & 0.603 \\
        & Mirror2-GS (Pose Cond.)
        & \second{0.192} & \second{0.232}
        & \third{0.231} & 0.229
        & \second{0.239} & \third{0.174}
        & \second{0.306} & 0.348
        & \second{0.264} & \second{0.515}
        & \second{0.180} & 0.044
        & \second{0.314} & \second{0.502} \\
        & \textbf{Ours (Pose Cond.)}
        & \first{0.181} & \first{0.214}
        & \first{0.185} & \first{0.164}
        & \first{0.212} & \first{0.143}
        & \first{0.251} & \second{0.304}
        & \first{0.212} & \first{0.447}
        & \first{0.114} & \second{0.031}
        & \first{0.172} & \first{0.318} \\
        \bottomrule
    \end{tabular}%
    }%
    }
    \vspace{-0.15in}
\end{table*}

\begin{table*}[t]
    \centering
    \caption{Quantitative comparison across seven benchmark datasets using \textbf{32 input views}. The top-3 results are highlighted as \colorbox{rankfirst}{first}, \colorbox{ranksecond}{second}, and \colorbox{rankthird}{third}.}
    \label{tab:main_comparison_32view}
    \setlength{\tabcolsep}{1pt}
    \renewcommand{\arraystretch}{1.2}
    \small
    \makebox[\textwidth][c]{%
    \resizebox{\textwidth}{!}{%
    \begin{tabular}{clcccccccccccccc}
        \toprule
        \multirow{2}{*}{\textbf{Input}}
        & \multirow{2}{*}{\textbf{Method}}
        & \multicolumn{2}{c}{\textbf{ScanNet}}
        & \multicolumn{2}{c}{\textbf{NRGBD}}
        & \multicolumn{2}{c}{\textbf{DL3DV}}
        & \multicolumn{2}{c}{\textbf{Tanks \& Temples}}
        & \multicolumn{2}{c}{\textbf{Mip-NeRF 360}}
        & \multicolumn{2}{c}{\textbf{ETH3D}}
        & \multicolumn{2}{c}{\textbf{DTU}} \\
        \cmidrule(lr){3-4}
        \cmidrule(lr){5-6}
        \cmidrule(lr){7-8}
        \cmidrule(lr){9-10}
        \cmidrule(lr){11-12}
        \cmidrule(lr){13-14}
        \cmidrule(lr){15-16}
        & & DISTS$\downarrow$ & LPIPS$\downarrow$
        & DISTS$\downarrow$ & LPIPS$\downarrow$
        & DISTS$\downarrow$ & LPIPS$\downarrow$
        & DISTS$\downarrow$ & LPIPS$\downarrow$
        & DISTS$\downarrow$ & LPIPS$\downarrow$
        & DISTS$\downarrow$ & LPIPS$\downarrow$
        & DISTS$\downarrow$ & LPIPS$\downarrow$ \\
        \midrule
        \multirow{6}{*}{\rotatebox[origin=c]{90}{w/o Pose}}
        & AnySplat
        & 0.300 & 0.430
        & \second{0.232} & 0.342
        & 0.380 & 0.338
        & 0.655 & 0.811
        & 0.618 & 0.833
        & 0.241 & 0.106
        & 0.577 & 0.805 \\
        & YoNoSplat
        & 0.268 & \third{0.238}
        & 0.296 & 0.253
        & 0.382 & \third{0.225}
        & 0.430 & \second{0.342}
        & 0.400 & 0.581
        & 0.316 & \first{0.039}
        & 0.497 & 0.706 \\
        & DA3-GS
        & 0.270 & 0.282
        & 0.249 & \third{0.233}
        & 0.355 & 0.234
        & \third{0.379} & 0.358
        & \third{0.352} & \third{0.557}
        & \third{0.211} & 0.091
        & \second{0.314} & \second{0.462} \\
        & Mirror2-GS
        & \second{0.192} & \second{0.229}
        & \third{0.238} & 0.247
        & \second{0.249} & \second{0.195}
        & \second{0.312} & \third{0.352}
        & \second{0.288} & \second{0.544}
        & \second{0.204} & 0.082
        & \third{0.371} & \third{0.623} \\
        & QuerySplat
        & \third{0.231} & 0.253
        & 0.263 & \second{0.223}
        & \third{0.343} & 0.301
        & 0.428 & 0.400
        & 0.408 & 0.658
        & 0.286 & \second{0.043}
        & 0.403 & 0.627 \\
        & \textbf{Ours}
        & \first{0.181} & \first{0.205}
        & \first{0.199} & \first{0.182}
        & \first{0.223} & \first{0.154}
        & \first{0.253} & \first{0.300}
        & \first{0.231} & \first{0.460}
        & \first{0.161} & \third{0.071}
        & \first{0.234} & \first{0.455} \\
        \midrule
        \multirow{7}{*}{\rotatebox[origin=c]{90}{w/ Pose}}
        & DepthSplat
        & \third{0.238} & 0.265
        & 0.313 & 0.277
        & \third{0.316} & 0.189
        & \third{0.340} & \third{0.341}
        & 0.530 & 0.646
        & 0.232 & \first{0.035}
        & 0.373 & 0.560 \\
        & TokenGS
        & 0.374 & 0.277
        & 0.332 & 0.282
        & 0.378 & 0.213
        & 0.534 & 0.354
        & 0.519 & 0.740
        & 0.347 & \second{0.043}
        & 0.554 & 0.772 \\
        & tttLRM
        & 0.281 & 0.286
        & 0.292 & 0.250
        & \third{0.316} & 0.190
        & 0.386 & 0.346
        & 0.334 & 0.525
        & \third{0.199} & 0.066
        & 0.349 & 0.528 \\
        & ReSplat
        & 0.271 & \third{0.245}
        & 0.302 & 0.249
        & 0.328 & \second{0.170}
        & 0.368 & \first{0.263}
        & \third{0.323} & \second{0.517}
        & 0.242 & \first{0.035}
        & 0.343 & \third{0.512} \\
        & DA3-GS (Pose Cond.)
        & 0.254 & 0.259
        & \third{0.255} & \third{0.244}
        & 0.336 & 0.235
        & 0.397 & 0.354
        & 0.340 & 0.537
        & 0.202 & 0.088
        & \third{0.324} & \second{0.473} \\
        & Mirror2-GS (Pose Cond.)
        & \second{0.194} & \second{0.230}
        & \second{0.237} & \second{0.241}
        & \second{0.241} & \third{0.185}
        & \second{0.309} & 0.346
        & \second{0.277} & \third{0.523}
        & \second{0.193} & 0.070
        & \second{0.322} & 0.521 \\
        & \textbf{Ours (Pose Cond.)}
        & \first{0.184} & \first{0.212}
        & \first{0.186} & \first{0.161}
        & \first{0.213} & \first{0.147}
        & \first{0.254} & \second{0.303}
        & \first{0.219} & \first{0.452}
        & \first{0.139} & \third{0.057}
        & \first{0.185} & \first{0.357} \\
        \bottomrule
    \end{tabular}%
    }%
    }
    \vspace{-0.15in}
    \par\vspace{3pt}
\centering
\caption{Quantitative comparison across seven benchmark datasets using \textbf{64 input views}. ETH3D and DTU results are unavailable and denoted by ``--''. The top-3 available results are highlighted as \colorbox{rankfirst}{first}, \colorbox{ranksecond}{second}, and \colorbox{rankthird}{third}.}
\label{tab:main_comparison_64view}
\setlength{\tabcolsep}{1pt}
\renewcommand{\arraystretch}{1.2}
\small
    \makebox[\textwidth][c]{%
    \resizebox{\textwidth}{!}{%
    \begin{tabular}{clcccccccccccccc}
        \toprule
        \multirow{2}{*}{\textbf{Input}}
        & \multirow{2}{*}{\textbf{Method}}
        & \multicolumn{2}{c}{\textbf{ScanNet}}
        & \multicolumn{2}{c}{\textbf{NRGBD}}
        & \multicolumn{2}{c}{\textbf{DL3DV}}
        & \multicolumn{2}{c}{\textbf{Tanks \& Temples}}
        & \multicolumn{2}{c}{\textbf{Mip-NeRF 360}}
        & \multicolumn{2}{c}{\textbf{ETH3D}}
        & \multicolumn{2}{c}{\textbf{DTU}} \\
        \cmidrule(lr){3-4}
        \cmidrule(lr){5-6}
        \cmidrule(lr){7-8}
        \cmidrule(lr){9-10}
        \cmidrule(lr){11-12}
        \cmidrule(lr){13-14}
        \cmidrule(lr){15-16}
        & & DISTS$\downarrow$ & LPIPS$\downarrow$
        & DISTS$\downarrow$ & LPIPS$\downarrow$
        & DISTS$\downarrow$ & LPIPS$\downarrow$
        & DISTS$\downarrow$ & LPIPS$\downarrow$
        & DISTS$\downarrow$ & LPIPS$\downarrow$
        & DISTS$\downarrow$ & LPIPS$\downarrow$
        & DISTS$\downarrow$ & LPIPS$\downarrow$ \\
        \midrule
        \multirow{6}{*}{\rotatebox[origin=c]{90}{w/o Pose}}
        & AnySplat
        & \third{0.215} & 0.287
        & \third{0.227} & 0.337
        & 0.393 & 0.333
        & 0.397 & 0.483
        & 0.388 & 0.658
        & -- & --
        & -- & -- \\
        & YoNoSplat
        & 0.272 & 0.237
        & 0.292 & 0.261
        & 0.373 & \third{0.215}
        & 0.413 & \third{0.321}
        & 0.422 & 0.608
        & -- & --
        & -- & -- \\
        & DA3-GS
        & 0.268 & 0.268
        & \second{0.211} & \second{0.201}
        & 0.339 & \third{0.231}
        & \third{0.343} & \second{0.311}
        & \third{0.331} & \third{0.512}
        & -- & --
        & -- & -- \\
        & Mirror2-GS
        & \second{0.205} & \second{0.208}
        & 0.236 & 0.250
        & \second{0.249} & \second{0.194}
        & \second{0.315} & 0.346
        & \second{0.305} & \second{0.495}
        & -- & --
        & -- & -- \\
        & QuerySplat
        & 0.237 & \third{0.228}
        & 0.261 & \third{0.223}
        & \third{0.334} & 0.274
        & 0.400 & 0.370
        & 0.390 & 0.634
        & -- & --
        & -- & -- \\
        & \textbf{Ours}
        & \first{0.190} & \first{0.182}
        & \first{0.187} & \first{0.183}
        & \first{0.222} & \first{0.161}
        & \first{0.271} & \first{0.271}
        & \first{0.248} & \first{0.434}
        & -- & --
        & -- & -- \\
        \midrule
        \multirow{7}{*}{\rotatebox[origin=c]{90}{w/ Pose}}
        & DepthSplat
        & \third{0.249} & 0.256
        & 0.313 & 0.335
        & 0.349 & 0.219
        & 0.359 & 0.355
        & 0.489 & 0.604
        & -- & --
        & -- & -- \\
        & TokenGS
        & 0.294 & \third{0.239}
        & 0.323 & 0.281
        & 0.368 & 0.223
        & 0.445 & 0.314
        & 0.444 & 0.671
        & -- & --
        & -- & -- \\
        & tttLRM
        & 0.280 & 0.270
        & 0.276 & \third{0.238}
        & 0.323 & 0.193
        & 0.375 & 0.336
        & 0.322 & \third{0.451}
        & -- & --
        & -- & -- \\
        & ReSplat
        & 0.287 & 0.250
        & 0.348 & 0.336
        & 0.329 & \second{0.173}
        & 0.352 & \first{0.244}
        & \third{0.292} & \second{0.429}
        & -- & --
        & -- & -- \\
        & DA3-GS (Pose Cond.)
        & 0.253 & 0.247
        & \second{0.213} & \second{0.208}
        & \third{0.306} & 0.213
        & \third{0.342} & \third{0.312}
        & 0.313 & 0.481
        & -- & --
        & -- & -- \\
        & Mirror2-GS (Pose Cond.)
        & \second{0.208} & \second{0.205}
        & \third{0.233} & \third{0.238}
        & \second{0.242} & \third{0.188}
        & \second{0.316} & 0.340
        & \second{0.286} & 0.463
        & -- & --
        & -- & -- \\
        & \textbf{Ours (Pose Cond.)}
        & \first{0.191} & \first{0.177}
        & \first{0.182} & \first{0.160}
        & \first{0.217} & \first{0.155}
        & \first{0.267} & \second{0.256}
        & \first{0.228} & \first{0.395}
        & -- & --
        & -- & -- \\
        \bottomrule
    \end{tabular}%
    }%
    }
    \vspace{-0.15in}

    \par\vspace{2pt}
    \centering
    \caption{Average PSNR comparison under different numbers of input views. Results are averaged across the different datasets. The top-3 results are highlighted as \colorbox{rankfirst}{first}, \colorbox{ranksecond}{second}, and \colorbox{rankthird}{third}.}
    \vspace{0.05in}
    \label{tab:main_comparison_psnr}
    \setlength{\tabcolsep}{8.5pt}
    \renewcommand{\arraystretch}{1}
    \small
    \begin{tabular}{clccccc}
        \toprule
        \multirow{2}{*}{\textbf{Input}}
        & \multirow{2}{*}{\textbf{Method}}
        & \textbf{4 Views}
        & \textbf{8 Views}
        & \textbf{16 Views}
        & \textbf{32 Views}
        & \textbf{64 Views} \\
        & & PSNR$\uparrow$ & PSNR$\uparrow$ & PSNR$\uparrow$ & PSNR$\uparrow$ & PSNR$\uparrow$ \\
        \midrule
        \multirow{6}{*}{\rotatebox[origin=c]{90}{w/o Pose}}
        & AnySplat
        & 12.68 & 13.20 & 12.83 & 13.12 & 14.26 \\
        & YoNoSplat
        & \first{18.23} & \first{18.64} & \first{18.73} & \first{18.34} & 15.79 \\
        & DA3-GS
        & 16.32 & 16.94 & 16.73 & 17.17 & \second{17.26} \\
        & Mirror2-GS
        & 17.20 & 17.60 & 17.98 & 17.04 & \third{17.09} \\
        & QuerySplat
        & \third{17.77} & \third{18.26} & \third{18.33} & \third{17.84} & 16.90 \\
        & \textbf{Ours}
        & \second{17.85} & \second{18.28} & \second{18.61} & \second{17.85} & \first{17.98} \\
        \midrule
        \multirow{7}{*}{\rotatebox[origin=c]{90}{w/ Pose}}
        & DepthSplat
        & 15.60 & 16.89 & 17.24 & 17.68 & 16.35 \\
        & TokenGS
        & \third{17.65} & \third{18.42} & \third{18.77} & \third{18.33} & \third{17.95} \\
        & tttLRM
        & 15.35 & 16.68 & 17.61 & 17.13 & 16.94 \\
        & ReSplat
        & \first{19.03} & \first{19.52} & \first{19.67} & \first{19.44} & \second{18.23} \\
        & DA3-GS (Pose Cond.)
        & 16.18 & 16.96 & 16.99 & 17.32 & 17.55 \\
        & Mirror2-GS (Pose Cond.)
        & 17.53 & 18.00 & 18.36 & 17.56 & 17.43 \\
        & \textbf{Ours (Pose Cond.)}
        & \second{18.39} & \second{18.82} & \second{19.12} & \second{18.39} & \first{18.50} \\
        \bottomrule
    \end{tabular}
\vspace{-0.15in}

\end{table*}

\subsection{Additional Quantitative Results}

\label{sec:app_n_view}

\paragraph{More quantitative comparisons.}
We report the full comparisons using \textbf{4}, \textbf{8}, \textbf{16}, \textbf{32}, and \textbf{64} input views, as shown in Table~\ref{tab:main_comparison_4view}, Table~\ref{tab:main_comparison_8view}, Table~\ref{tab:main_comparison_16view}, Table~\ref{tab:main_comparison_32view}, and Table~\ref{tab:main_comparison_64view}. 

\paragraph{PSNR metric results.}
Table~\ref{tab:main_comparison_psnr} reports the average PSNR of different methods. However, PSNR does not always faithfully reflect perceptual rendering quality, as it can favor overly smooth predictions that achieve low pixel-wise errors while losing high-frequency details. Although ReSplat achieves the highest PSNR in several settings, its extrapolation capability remains limited. Under large viewpoint changes, the reconstructed Gaussians often produce severe artifacts, leading to noticeable degradation in rendering quality, as shown in Fig.~\ref{fig:qualitative} and Fig.~\ref{fig:appendix_comparisons}. In contrast, our method delivers substantially higher rendering quality under extrapolated viewpoints.

\vspace{-0.1in}
\subsection{Additional Qualitative Results}

\begin{figure*}[t] 
    \centering
    \includegraphics[width=\textwidth]{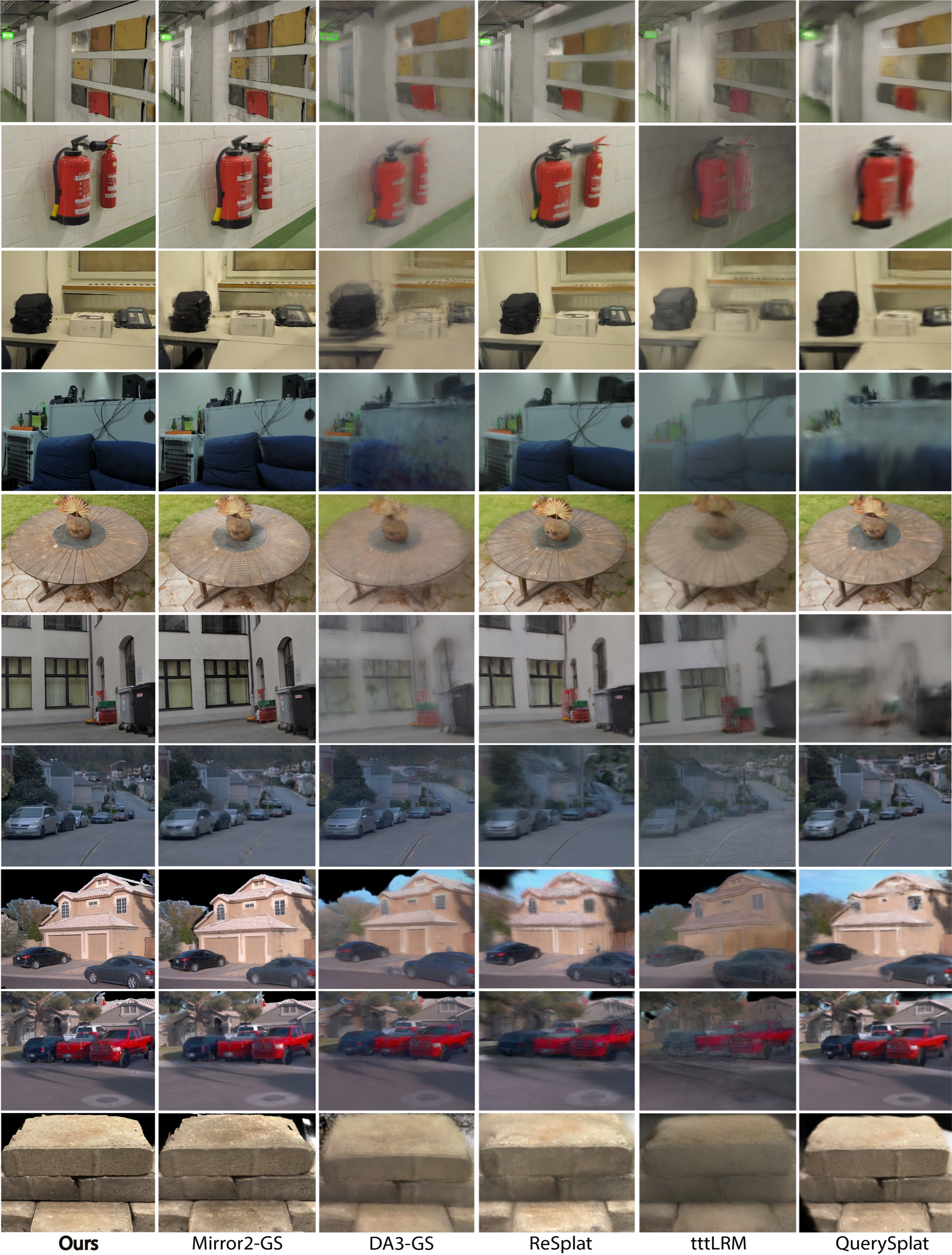}
    \vspace{-0.25in}
    \caption{Qualitative comparisons across diverse scenes and viewpoints with multi-view input.}
    \label{fig:appendix_comparisons}
    \vspace{-0.2in}
\end{figure*}

\begin{figure*}[t] 
    \centering
    \includegraphics[width=\textwidth]{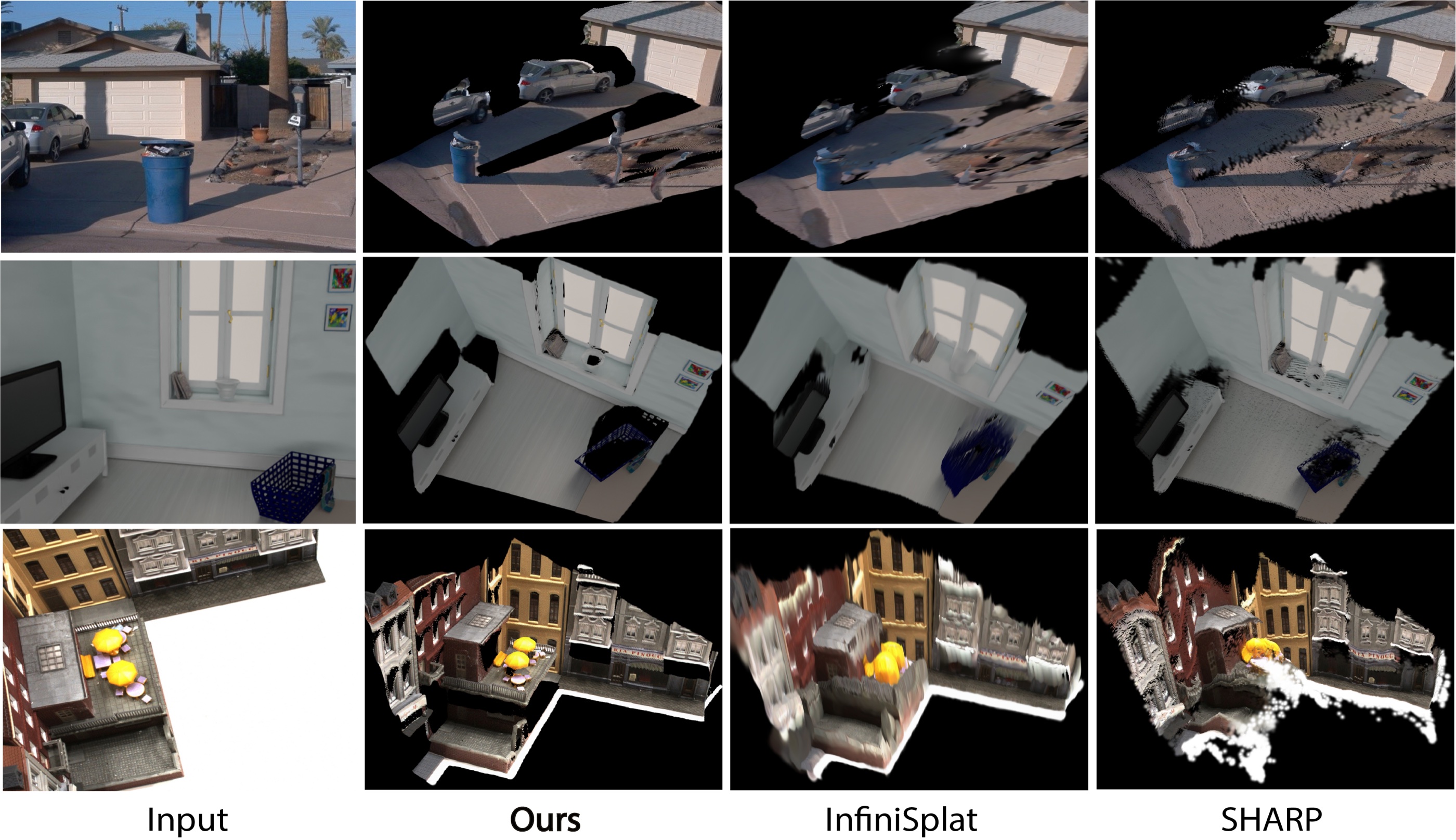}
    \vspace{-0.2in}
    \caption{Qualitative comparisons across diverse scenes and viewpoints with monocular input.}
    \label{fig:appendix_comparisons_mono}
    
\end{figure*}

\paragraph{More qualitative comparisons.}
We provide more comparisons of SplatAnything with other methods under multi-view and monocular input settings. Fig.~\ref{fig:appendix_comparisons} and Fig.~\ref{fig:appendix_comparisons_mono} demonstrate the performance of SplatAnything under multi-view and monocular input settings, respectively.
SplatAnything demonstrates superior performance over other methods under large viewpoint shifts.

\begin{figure*}[t] 
    \centering
    \includegraphics[width=\textwidth]{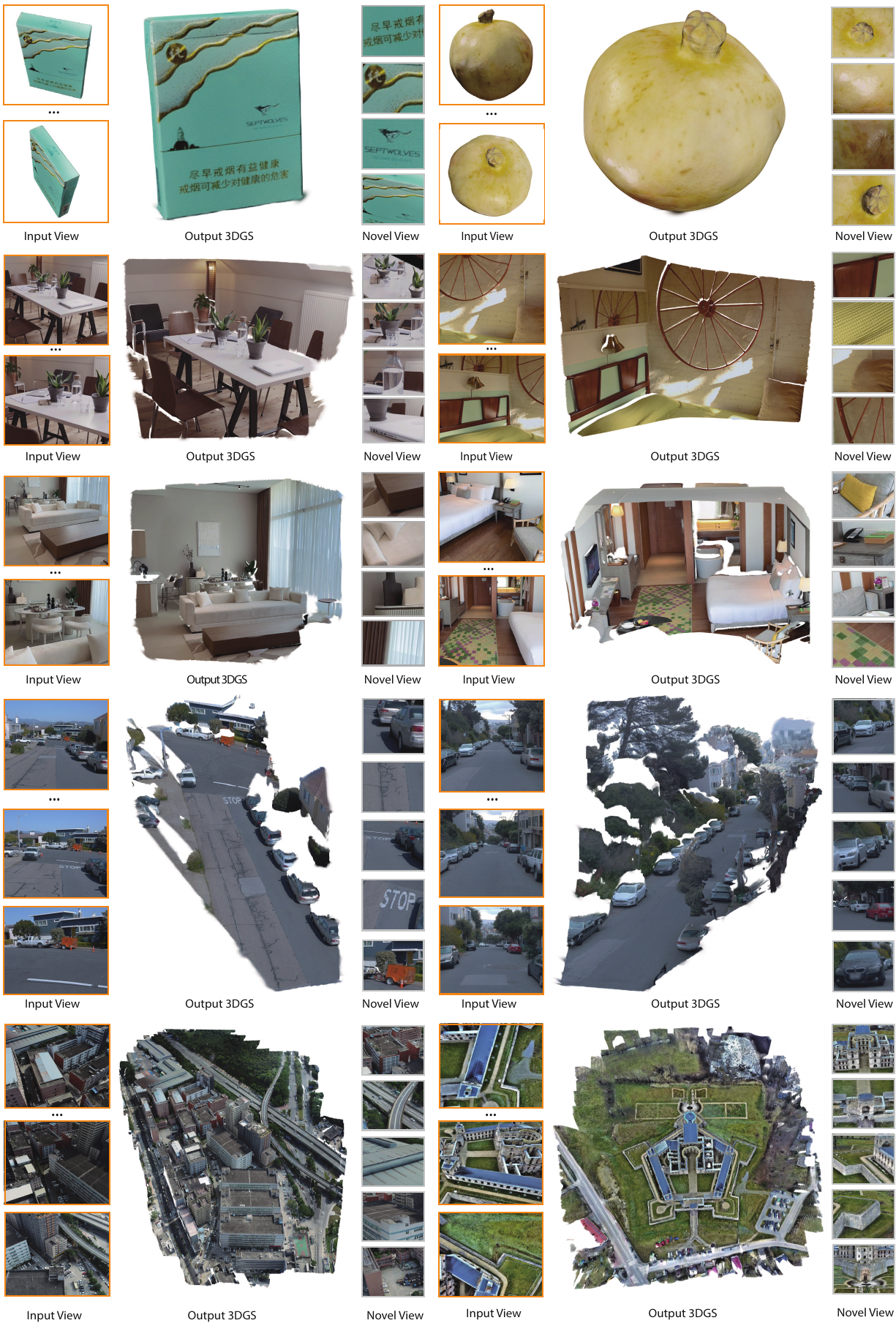}
    \vspace{-0.2in}
    \caption{More qualitative results of our method. SplatAnything feed-forward reconstructs high-quality 3DGS representations generalizing to arbitrary scenes, scales, and viewpoints in a few seconds.}
    \label{fig:more_qualitative_results}
\end{figure*}

\paragraph{More qualitative results of SplatAnything.}
We provide more qualitative results of SplatAnything, including objects, indoor scenes, and large-scale outdoor scenes. Results are shown in Fig.~\ref{fig:more_qualitative_results}. 

\begin{table*}[t]
    \centering
    \caption{Median forward-pass latency per scene (ms) of different methods under different numbers of input views. The fastest method within each pose setting is highlighted in bold.}
    \vspace{0.1in}
    \label{tab:efficiency}
    \setlength{\tabcolsep}{9pt}
    \renewcommand{\arraystretch}{1.0}
    \small
    \begin{tabular}{clccccc}
        \toprule
        \textbf{Input} & \textbf{Method}
        & \textbf{$N=4$}
        & \textbf{$N=8$}
        & \textbf{$N=16$}
        & \textbf{$N=32$}
        & \textbf{$N=64$} \\
        \midrule
        \multirow{5}{*}{\rotatebox[origin=c]{90}{w/o Pose}}
        & AnySplat
        & 250 & 487 & 1,089 & 2,743 & 7,798 \\
        & DA3-GS
        & 709 & 1,513 & 3,783 & 10,950 & 35,904 \\
        & Mirror2-GS
        & \textbf{246} & \textbf{448} & \textbf{959} & \textbf{2,268} & \textbf{6,060} \\
        & QuerySplat
        & 2,324 & 3,398 & 6,092 & 14,009 & 39,583 \\
        & \textbf{Ours}
        & 411 & 761 & 1,649 & 3,991 & 11,037 \\
        \midrule
        \multirow{7}{*}{\rotatebox[origin=c]{90}{w/ Pose}}
        & DepthSplat
        & \textbf{158} & \textbf{287} & \textbf{541} & \textbf{1,045} & \textbf{2,087} \\
        & TokenGS
        & 541 & 1,225 & 3,460 & 11,447 & 41,480 \\
        & tttLRM
        & 487 & 886 & 1,681 & 3,271 & 6,465 \\
        & ReSplat
        & 199 & 376 & 855 & 2,041 & 5,568 \\
        & DA3-GS (Pose Cond.)
        & 711 & 1,515 & 3,791 & 10,953 & 35,892 \\
        & Mirror2-GS (Pose Cond.)
        & 250 & 454 & 968 & 2,288 & 6,131 \\
        & \textbf{Ours (Pose Cond.)}
        & 413 & 764 & 1,651 & 4,007 & 11,045 \\
        \bottomrule
    \end{tabular}
\end{table*}

\begin{table*}[t]
{
\centering
\caption{Ablation study of different normalization spaces across different numbers of input views.}
\vspace{0.05in}
\label{tab:ablation_normalization_space_full}
\setlength{\tabcolsep}{1pt}
\renewcommand{\arraystretch}{1.2}
\small
\makebox[\textwidth][c]{%
\resizebox{\textwidth}{!}{%
    \begin{tabular}{lcccccccccccc}
        \toprule
        \multirow{2}{*}{\textbf{Norm. Space}}
        & \multicolumn{2}{c}{\textbf{Num. Views=1}}
        & \multicolumn{2}{c}{\textbf{Num. Views=4}}
        & \multicolumn{2}{c}{\textbf{Num. Views=8}}
        & \multicolumn{2}{c}{\textbf{Num. Views=16}}
        & \multicolumn{2}{c}{\textbf{Num. Views=32}}
        & \multicolumn{2}{c}{\textbf{Num. Views=64}} \\
        \cmidrule(lr){2-3}
        \cmidrule(lr){4-5}
        \cmidrule(lr){6-7}
        \cmidrule(lr){8-9}
        \cmidrule(lr){10-11}
        \cmidrule(lr){12-13}
        & DISTS$\downarrow$ & LPIPS$\downarrow$
        & DISTS$\downarrow$ & LPIPS$\downarrow$
        & DISTS$\downarrow$ & LPIPS$\downarrow$
        & DISTS$\downarrow$ & LPIPS$\downarrow$
        & DISTS$\downarrow$ & LPIPS$\downarrow$
        & DISTS$\downarrow$ & LPIPS$\downarrow$ \\
        \midrule
        Radial Distance Norm.
        & 0.308 & 0.394
        & 0.223 & 0.278
        & 0.223 & 0.272
        & 0.218 & 0.268
        & 0.225 & 0.277
        & 0.224 & 0.250 \\
        Camera Distance Norm.
        & 0.793 & 1.027
        & 0.217 & 0.284
        & 0.216 & 0.280
        & 0.214 & 0.280
        & 0.221 & 0.291
        & 0.224 & 0.260 \\
        Depth Norm.
        & 0.307 & 0.413
        & 0.224 & 0.284
        & 0.225 & 0.278
        & 0.218 & 0.273
        & 0.224 & 0.282
        & 0.224 & 0.254 \\
        \textbf{Gaussian Spacing Norm. (Ours)}
        & \textbf{0.294} & \textbf{0.367}
        & \textbf{0.198} & \textbf{0.245}
        & \textbf{0.196} & \textbf{0.237}
        & \textbf{0.190} & \textbf{0.232}
        & \textbf{0.197} & \textbf{0.241}
        & \textbf{0.217} & \textbf{0.229} \\
        \bottomrule
    \end{tabular}%
}%
}
\par}
\vspace{3pt}
{
\centering
\vspace{-0.1in}
\caption{Ablation study of the training design choices across different numbers of input views.}
\vspace{0.05in}
\label{tab:ablation_design_choices}
\setlength{\tabcolsep}{0.9pt}
\renewcommand{\arraystretch}{1.3}
\small
\makebox[\textwidth][c]{%
\resizebox{\textwidth}{!}{%
    \begin{tabular}{lcccccccccccc}
        \toprule
        \multirow{2}{*}{\textbf{Methods}}
        & \multicolumn{2}{c}{\textbf{Num. Views=1}}
        & \multicolumn{2}{c}{\textbf{Num. Views=4}}
        & \multicolumn{2}{c}{\textbf{Num. Views=8}}
        & \multicolumn{2}{c}{\textbf{Num. Views=16}}
        & \multicolumn{2}{c}{\textbf{Num. Views=32}}
        & \multicolumn{2}{c}{\textbf{Num. Views=64}} \\
        \cmidrule(lr){2-3}
        \cmidrule(lr){4-5}
        \cmidrule(lr){6-7}
        \cmidrule(lr){8-9}
        \cmidrule(lr){10-11}
        \cmidrule(lr){12-13}
        & DISTS$\downarrow$ & LPIPS$\downarrow$
        & DISTS$\downarrow$ & LPIPS$\downarrow$
        & DISTS$\downarrow$ & LPIPS$\downarrow$
        & DISTS$\downarrow$ & LPIPS$\downarrow$
        & DISTS$\downarrow$ & LPIPS$\downarrow$
        & DISTS$\downarrow$ & LPIPS$\downarrow$ \\
        \midrule
        Ours w/o Fine-tuning Gaussian Backbone
        & 0.2952 & 0.3809
        & 0.2026 & 0.2503
        & 0.2006 & 0.2421
        & 0.1948 & 0.2379
        & 0.2038 & 0.2467
        & 0.2239 & 0.2387 \\
        Ours w/o Fixed Gaussian Centers
        & 0.2989 & 0.4010
        & 0.2063 & 0.2591
        & 0.2040 & 0.2512
        & 0.1981 & 0.2463
        & 0.2079 & 0.2584
        & 0.2259 & 0.2507 \\
        Ours w/o Data Scaling
        & 0.2948 & 0.3755
        & 0.2045 & 0.2499
        & 0.2027 & 0.2434
        & 0.1960 & 0.2389
        & 0.2041 & 0.2489
        & 0.2229 & 0.2385 \\
        Ours w/o View Sampling Strategy
        & 0.2974 & 0.3996
        & 0.2026 & 0.2536
        & 0.2001 & 0.2472
        & 0.1928 & 0.2392
        & 0.1995 & 0.2475
        & 0.2231 & 0.2448 \\
        \textbf{Ours}
        & \textbf{0.2937} & \textbf{0.3667}
        & \textbf{0.1981} & \textbf{0.2450}
        & \textbf{0.1956} & \textbf{0.2374}
        & \textbf{0.1895} & \textbf{0.2315}
        & \textbf{0.1970} & \textbf{0.2412}
        & \textbf{0.2170} & \textbf{0.2285} \\
        \bottomrule
    \end{tabular}%
    }%
    }
\par}
\vspace{10pt}
\begin{minipage}{\textwidth}
    \centering
    \includegraphics[width=\textwidth]{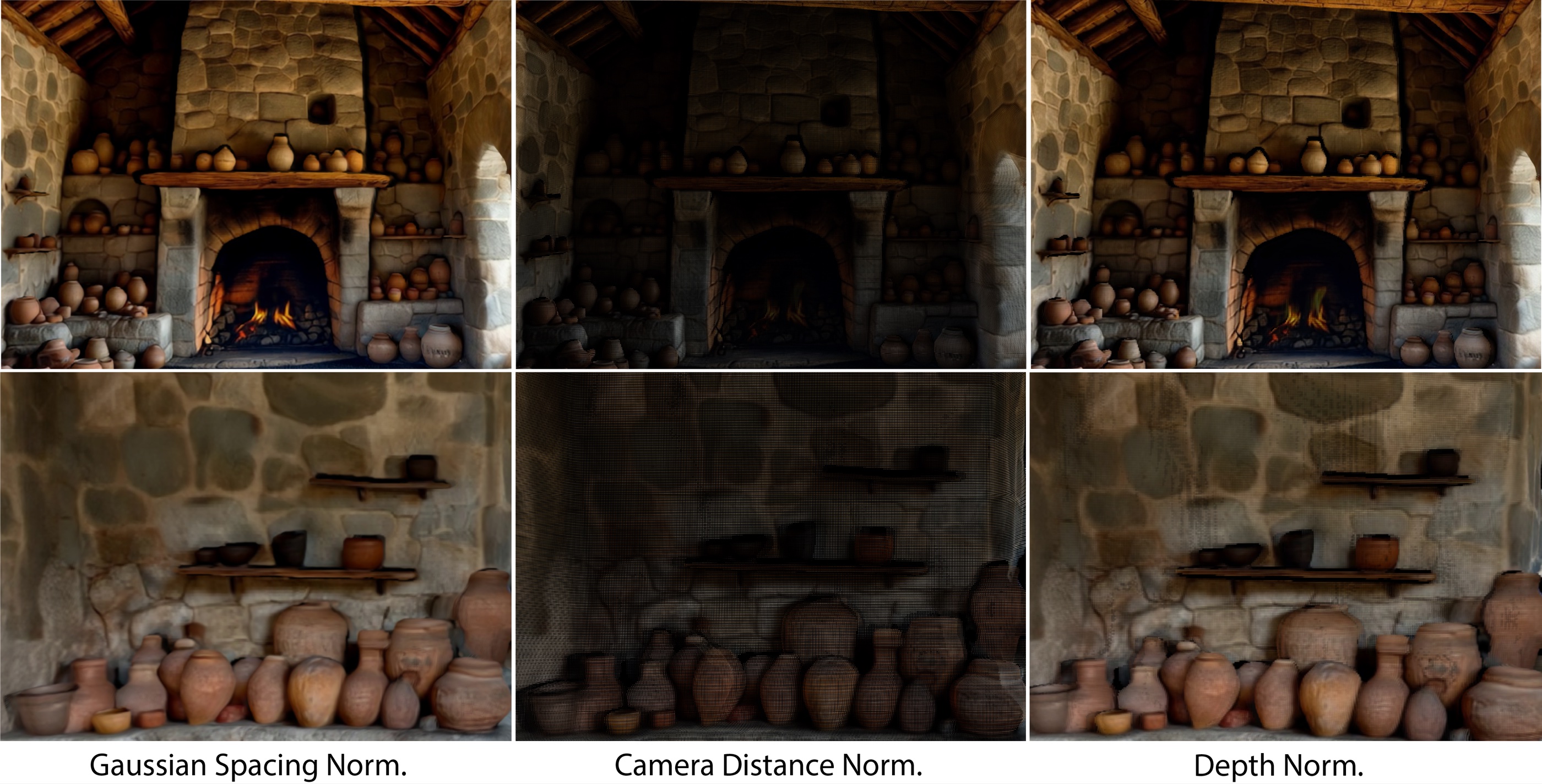}
    \vspace{-0.15in}
    \captionof{figure}{Qualitative comparisons of different normalization spaces.}
    \label{fig:ablation_normalization_space_full}
    \vspace{-0.2in}
\end{minipage}

\end{table*}

\subsection{Inference Efficiency}

\label{sec:app_efficiency}
We measure the median forward-pass latency per scene for producing Gaussian attributes under different numbers of input views. As shown in Table~\ref{tab:efficiency}, the runtime of all methods increases with the number of input views. SplatAnything remains substantially faster than computationally heavier baselines such as DA3-GS, QuerySplat, and TokenGS. All methods are infered on a single NVIDIA H20 GPU. Inference uses BF16 precision and a batch size of 1. All input images are resized to the same resolution of $448\times896 (H \times W)$ to match the input constraints of all methods.

\subsection{Additional Ablation Studies}

\paragraph{Normalization Space for Gaussian Prediction.}

The effect of Gaussian Spacing Normalization is evaluated in the main
paper. Here, we provide a more comprehensive comparison of different
normalization spaces, including \textit{Radial Distance Normalization
(RDNorm)}, \textit{Camera Distance Normalization (CDNorm)},
\textit{Depth Normalization (DNorm)}, and our proposed
\textit{Gaussian Spacing Normalization (GSNorm)}.
RDNorm is described in the main paper.
CDNorm uses the median distance from all camera centers to a reference
camera as the scene-level normalization factor, whereas DNorm uses the
median depth over all frames.
Quantitative results are reported in
Table~\ref{tab:ablation_normalization_space_full}.

CDNorm is highly sensitive to camera trajectory: a small camera baseline
yields a small normalization factor and excessively large Gaussian
spacing in the normalized space, whereas larger camera displacement
induces stronger rescaling.
DNorm is less sensitive to camera motion, but normalizes only by scene
depth and therefore ignores the pixel-space focal length and image
sampling that directly affect Gaussian spacing.
Consequently, changes in field of view or image resolution can still
induce substantial variation in Gaussian-scale targets. 

To further illustrate the failure modes of CDNorm and DNorm, we examine
a challenging scene with a small camera baseline and a wide field of
view ($\sim100^\circ$).
As shown in Fig.~\ref{fig:ablation_normalization_space_full}, both
methods produce excessively large Gaussian spacing, but for different
reasons: CDNorm is dominated by the small camera displacement, whereas
DNorm fails to account for the spacing increase induced by the wide
field of view.
This shifts the Gaussian-spacing distribution away from that seen during
training, causing the predicted Gaussian scales to be too small for
adequate surface coverage and resulting in severe holes.
In contrast, GSNorm canonicalizes a robust scene-level statistic of
Gaussian spacing, providing a more consistent spatial reference across
varying camera trajectories and intrinsics.

\begin{figure*}[t] 
    \centering
    \includegraphics[width=\textwidth]{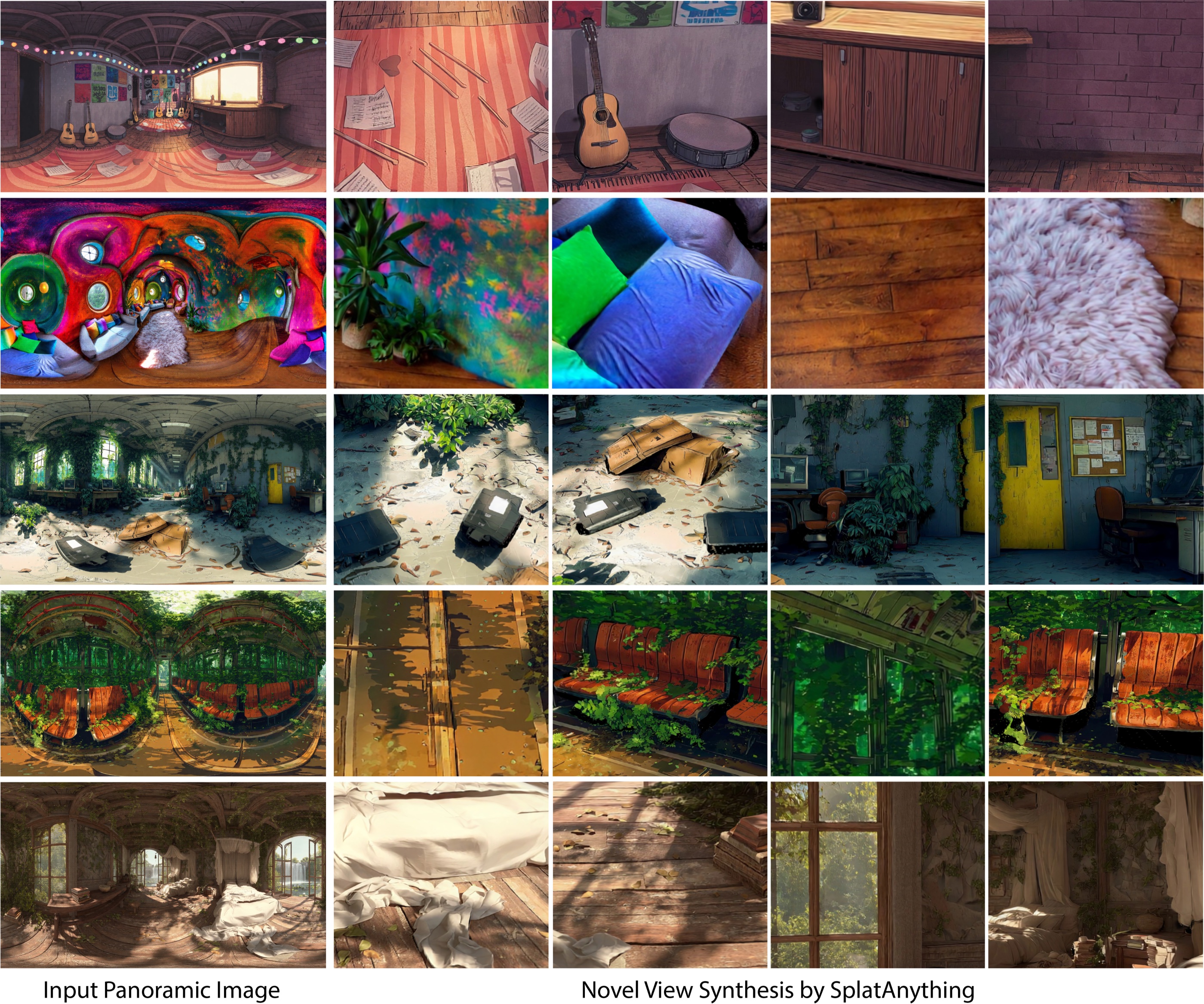}
    \vspace{-0.2in}
    \caption{Qualitative results of SplatAnything for panoramic images reconstruction. SplatAnything can reconstruct high-quality 3DGS representations from panoramic images with different field of view, demonstrating its strong generalization ability.}
    \label{fig:application}
\end{figure*}

\paragraph{Training Design Choices.}
We further ablate four design choices: fine-tuning the pretrained geometry backbone, fixing the Gaussian centers,
scaling the training data, and adopting diverse view-sampling strategies.
Results are reported in Table~\ref{tab:ablation_design_choices}.
All variants follow the same training settings. The upstream
WorldMirror~2.0 geometry estimator remains frozen in all experiments.

For \emph{w/o Fine-tuning Gaussian Backbone}, the Gaussian Backbone is
also frozen at its WorldMirror~2.0 initialization, and only GSHead is
optimized. For \emph{w/o Fixed Gaussian Centers}, GSHead additionally
predicts a per-pixel 3D center offset, instead of
fixing the centers to the unprojected geometry.
For \emph{w/o Data Scaling}, we train on Hypersim, while keeping the number of training iterations unchanged.
For \emph{w/o View Sampling Strategy}, we randomly select $n$ views and then partition them into source views and target views, while retaining the same numbers
of source and target views.
We find that fixing Gaussian centers produces the largest and most consistent gain across different input-view counts. When the centers are optimized only through rendering losses, the model can reduce image-space errors by
moving primitives away from the geometry prior, resulting in geometric
drift and poorer generalization to extrapolated viewpoints. Fine-tuning
the Gaussian Backbone adapts the pretrained geometric features to
appearance and Gaussian-attribute prediction, whereas scaling the
training data improves cross-domain robustness. The mixed view-sampling
strategy exposes the model to diverse camera motions and viewpoint
changes, leading to more reliable reconstruction under sparse inputs and
large extrapolation.

\subsection{Application to Panoramic 3D Content Generation.}
We explore SplatAnything for panoramic-image-based 3D
content generation.
We use WorldPano~\citep{team2026hy} to generate a panoramic image and
convert it into cubemap perspective views with a
$95^\circ \times 100^\circ$ FoV, which remains substantially wider than
typical perspective inputs.
These views are fed directly into SplatAnything for 3D Gaussian
reconstruction.
As shown in Fig.~\ref{fig:application}, SplatAnything produces
high-quality novel-view synthesis under this
challenging wide-FoV setting.
These results demonstrate its robustness to substantially different
camera configurations and further highlight the potential of
SplatAnything as a general feed-forward 3DGS framework.

\end{document}